\documentclass{article} 
\usepackage{graphicx}
\usepackage{iclr2027_conference,times}
\usepackage{caption}
\usepackage{subcaption}

\usepackage{amsmath,amsfonts,bm}

\def\eqref#1{equation~\ref{#1}}

\def\1{\bm{1}}

\DeclareMathAlphabet{\mathsfit}{\encodingdefault}{\sfdefault}{m}{sl}
\SetMathAlphabet{\mathsfit}{bold}{\encodingdefault}{\sfdefault}{bx}{n}

\usepackage{hyperref}
\usepackage{url}

\title{RoGSW4RLD: Feed-Forward 4D Gaussian \\
Lifting for Robot World Model Rollouts}

\author{
Jin Hyun Kim\textsuperscript{1},
Min Young Kim\textsuperscript{2},
Soohwan Song\textsuperscript{2}\footnotemark[1],
Daekyum Kim\textsuperscript{1,3}\thanks{
Corresponding authors: \texttt{daekyum@korea.ac.kr}; \texttt{songsh@dongguk.edu}.} \\
\textsuperscript{1}School of Mechanical Engineering, Korea University \\
\textsuperscript{2}College of AI Convergence, Dongguk University \\
\textsuperscript{3}School of Smart Mobility, Korea University \\
Seoul, Republic of Korea
}

\iclrfinalcopy
\begin{document}

\maketitle

\begin{figure}[h!]
\centering

\includegraphics[width=0.90\linewidth]{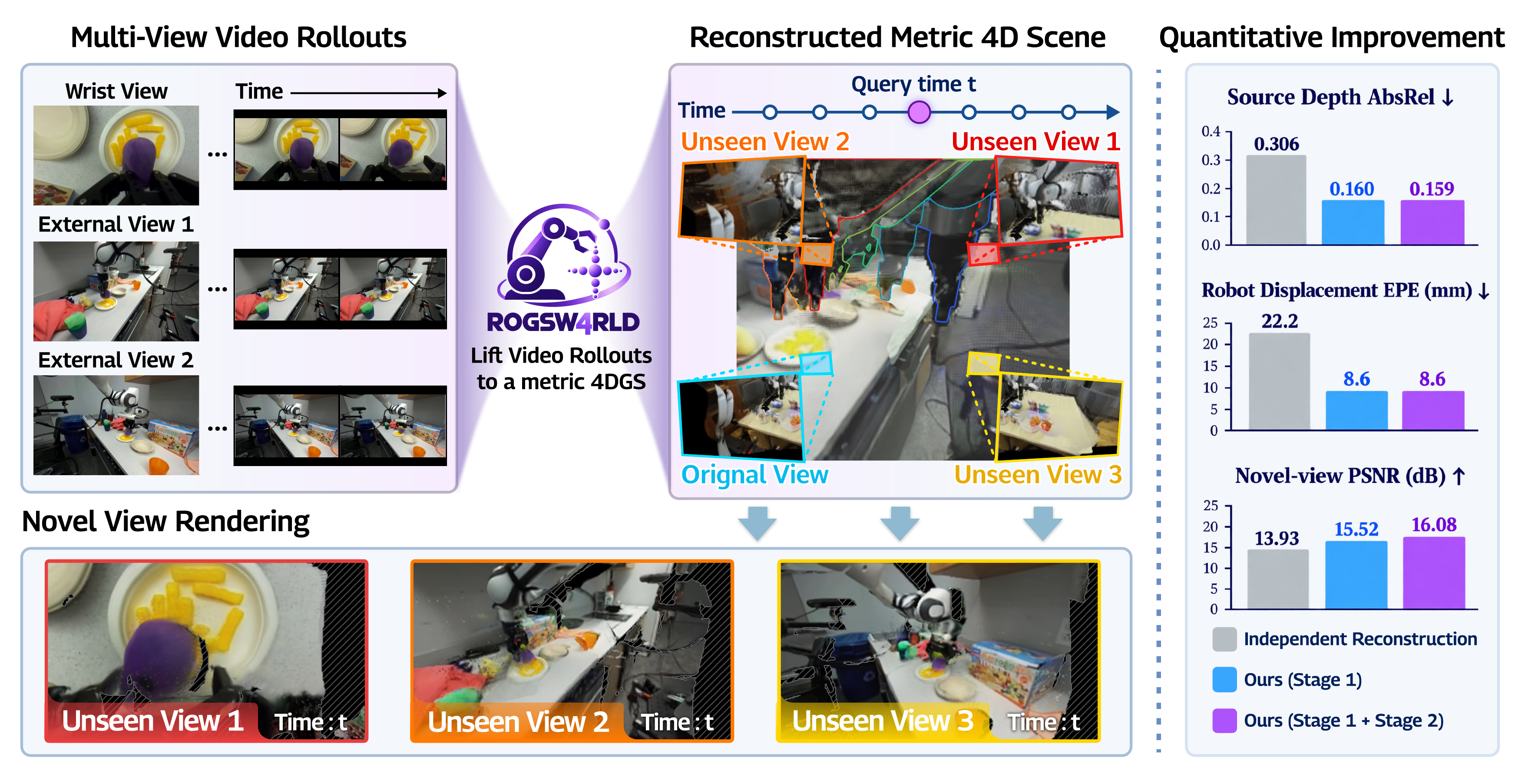}

\caption{
\textbf{Overview of RoGSW4RLD for metric 4D lifting of robot-camera rollouts.}
Given synchronized multi-view videos, RoGSW4RLD reconstructs a shared, time-queryable metric 4D Gaussian scene that supports rendering from unseen viewpoints at arbitrary query times.
Compared with independent camera-wise reconstruction using MoVieS~\citep{lin2026movies}, followed by stereo-depth metric normalization and calibrated merging in the robot frame, joint lifting substantially improves metric
depth, robot-motion accuracy, and novel-view rendering, with Stage~2 further
refining the reconstructed field.
Bars report results on recorded observations from
Table~\ref{tab:reconstruction_main}.
}
\label{fig:teaser}
\end{figure}

\begin{abstract}
Action-conditioned video world models predict future robot interactions from multiple cameras, yet their outputs remain disparate video collections rather than a shared metric scene queryable across viewpoints and time. While existing 4D reconstruction methods offer a path to spatialize these predictions, independently reconstructing and merging each camera stream fails to enforce cross-view consistency. This limitation is particularly detrimental when combining moving robot-mounted cameras with fixed external views. To address this, we introduce RoGSW4RLD, a feed-forward framework that lifts synchronized multi-camera rollouts into a unified, time-queryable metric 4D Gaussian field. Rather than learning a separate geometric transition model, RoGSW4RLD directly reconstructs the visual future generated by existing world models. Its core innovation is a two-stage architecture: Stage~1 jointly forms the metric 4D field by fusing cross-view evidence with robot-specific articulated geometry and kinematics, while Stage~2 refines the field's geometry and appearance while strictly preserving the initial temporal displacements. Evaluated on 256 held-out DROID episodes, RoGSW4RLD significantly outperforms camera-wise reconstruction with calibrated merging, improving novel-view PSNR by 2.15~dB, reducing depth AbsRel by 47\%, and lowering robot displacement error by 61\%. These robust gains extend to action-conditioned Cosmos~3 rollouts, demonstrating that predicted video futures can be successfully translated into consistent, spatially queryable 4D metric representations.

\end{abstract}

\section{Introduction}
\label{sec:introduction}

Video world models predict how physical interactions evolve under future actions~\citep{ebert2018visual,du2023unipi,yang2024unisim}. Spatial world models forecast geometric evolution~\citep{lu2025gwm,huang2026pointworld,zhen2025tesseract}, and related approaches construct dynamic Gaussian scenes from video generators~\citep{bahmani2026lyra,pan2026diff4splat}. In contrast, multi-camera models like Cosmos~3~\citep{nvidia2026cosmos3} output synchronized interaction videos but still lack a shared, temporally queryable metric scene. To address this, we lift these multi-view rollouts into a unified metric 4D Gaussian field (Figure~\ref{fig:teaser}). Rather than learning a new geometric transition model, our approach reconstructs the generated visual future directly.

\begin{figure}[t!]
\centering

\includegraphics[width=0.90\linewidth]{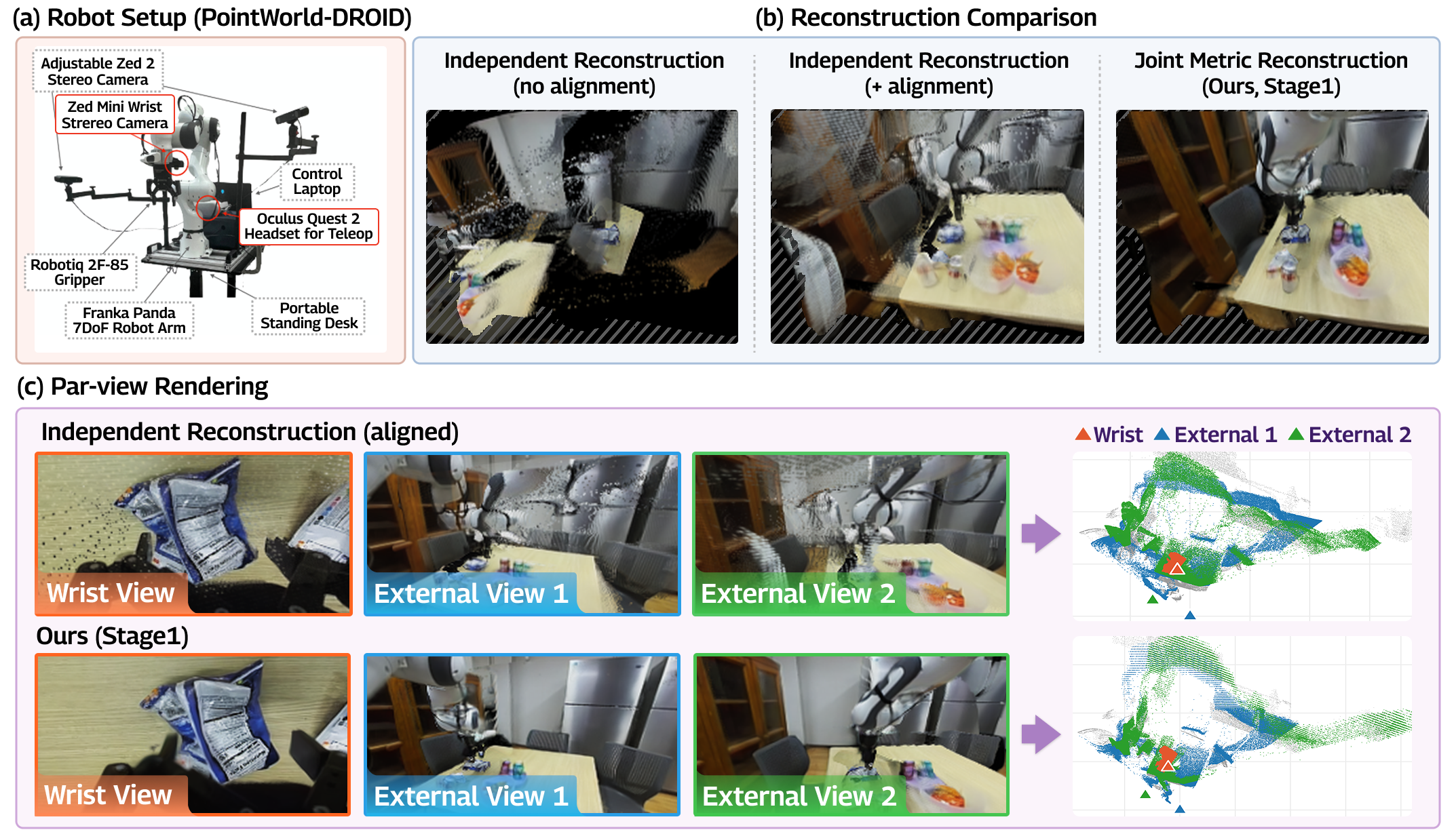}

\caption{
\textbf{Robot-camera setup and limitations of post-hoc field merging.}
(a) PointWorld-DROID setup with a moving wrist camera and fixed external cameras.
(b) Independent reconstruction retains inconsistent geometry after calibrated alignment, while our joint metric reconstruction (Stage~1) forms a more coherent scene.
(c) Per-view renderings and merged geometry show reduced duplication and misalignment across cameras.
}
\label{fig:setup_problem}
\end{figure}

A naive approach would independently reconstruct each camera stream and merge the fields using camera calibration. However, post-hoc alignment fails to enforce cross-camera consistency during geometry and motion estimation, often leaving duplicated or conflicting surfaces (Figure~\ref{fig:setup_problem}). This issue is especially prominent in setups like DROID~\citep{khazatsky2024droid}, which combine fixed external and moving robot-mounted cameras where views entangle ego-motion with scene dynamics. While recent feed-forward methods enable spatio-temporal reconstruction~\citep{ma20254dlrm,balice2026nopo4d}, robot-camera settings specifically require joint metric estimation guided by the robot's physical geometry and kinematics.

We propose RoGSW4RLD, a feed-forward framework that jointly lifts synchronized robot-camera rollouts into a shared metric 4D Gaussian field. Instead of relying on post-hoc alignment, our key insight is to integrate robot-specific spatial and motion priors directly during field formation. Using joint states, hand--eye calibration, articulated meshes, and forward kinematics (FK), we ground the geometry and motion estimation within a common robot frame, while inferring object and scene dynamics solely from the visual rollout.

Our method operates in two feed-forward stages without per-scene optimization. Stage~1 performs robot-aware lifting: combining multi-view evidence with articulated geometry to anchor robot surfaces at source times, while using FK to guide, rather than analytically replace, the learned deformations of all trajectories. Stage~2 applies displacement-preserving refinement to correct residual placement and appearance errors via source-view rendering feedback. By sharing position corrections across query times, this stage perfectly preserves the relative temporal displacements established in Stage~1.

Experiments on held-out DROID episodes demonstrate that RoGSW4RLD significantly improves novel-view rendering, metric geometry, and robot-motion accuracy compared to calibrated unions of independent reconstructions. These gains persist for action-conditioned Cosmos~3 rollouts, yielding a more consistent spatial representation of predicted futures.

Our contributions are threefold: i) a novel framework for jointly lifting synchronized multi-view robot rollouts into a shared metric 4D Gaussian field; ii) a two-stage architecture featuring robot-aware lifting and displacement-preserving refinement via articulated geometry and kinematics; and iii) substantial empirical improvements in metric geometry, novel-view synthesis, and motion accuracy on both recorded and generated robotic rollouts.

\section{Related Work}
\subsection{Video and Spatial World Models}

\paragraph{Video world models.}
Video prediction has evolved from action-conditioned dynamics for control~\citep{finn2016unsupervised,ebert2018visual} to comprehensive video-based planning and simulation~\citep{du2023unipi,yang2024unisim}. Recent models like Cosmos~3~\citep{nvidia2026cosmos3} and Ctrl-World~\citep{guo2026ctrlworld} generate controllable robot rollouts, while TesserAct~\citep{zhen2025tesseract} predicts RGB alongside depth and normals. Other approaches focus on improving cross-view consistency~\citep{liu2025geometryaware4d,wang2026mvista4d} or utilize rendered robot motion as a visual condition~\citep{kim2026robotfactored,gu2026geniworld}. Instead of modifying the video generation process itself, RoGSW4RLD focuses on lifting the synchronized rollouts of a frozen world model into a shared metric scene.

\paragraph{Spatial world models.}
Spatial world models directly forecast geometric evolution. For instance, GWM~\citep{lu2025gwm} learns action-conditioned transitions within a Gaussian latent space, and PointWorld~\citep{huang2026pointworld} predicts 3D point flows driven by robot surface motion. Rather than learning a novel geometric transition model, RoGSW4RLD leverages the supplied visual rollout as a predicted future, reconstructing its underlying geometry and motion into a unified metric 4D field.

\subsection{Dynamic Scene Reconstruction}

\paragraph{Optimization-based reconstruction.}
Dynamic neural and Gaussian fields typically model scene evolution via deformation or structured space-time representations~\citep{pumarola2021dnerf,park2021nerfies,wu2024fourDGS,yang2024deformable3dgs,lei2025mosca,wang2025shapeofmotion}. While these methods conventionally require per-scene optimization at test time, RoGSW4RLD employs a feed-forward approach, performing joint lifting and recurrent refinement using fixed network weights.

\paragraph{Feed-forward reconstruction.}
Feed-forward techniques amortize the reconstruction cost across scenes. For monocular videos, 4DGT~\citep{xu20254dgt} predicts temporally coherent Gaussians, and MoVieS~\citep{lin2026movies} estimates depth, Gaussian attributes, and query-time deformation. 4D-LRM~\citep{ma20254dlrm} extends reconstruction across varying views and times. Recent pose-free methods address monocular~\citep{yang2026neoverse}, driving~\citep{chen2025dggt}, two-image~\citep{hur2026ufo4d}, and multi-view~\citep{balice2026nopo4d} setups. Building upon MoVieS, our framework integrates robot geometry and kinematics into the joint metric reconstruction of calibrated moving or fixed cameras. Furthermore, we adapt ReSplat~\citep{xu2026resplat} by introducing a time-shared refinement mechanism that strictly preserves the per-Gaussian center displacements.

\paragraph{Generative priors for 4D reconstruction.}
Video generation models offer rich observations and features for reconstruction. CAT4D~\citep{wu2025cat4d} generates multi-view videos prior to optimizing dynamic Gaussians, whereas Diff4Splat~\citep{pan2026diff4splat} decodes video diffusion representations directly into dynamic Gaussians. Lyra~\citep{bahmani2026lyra} distills camera-controlled video diffusion into a feed-forward Gaussian decoder, extending this to dynamic video-to-4D generation. In contrast, we specifically target the joint metric lifting of robot-camera rollouts. We primarily rely on decoded RGB inputs, though an optional latent adapter can replace the token-input pathway for tighter integration.

\section{Problem Formulation}

We reconstruct dynamic robotic scenes from synchronized multi-view
observations, either recorded or generated by a video world model.
For source cameras $\mathcal V$ and source times $\mathcal S$, inputs comprise
RGB images $\mathbf I_{v,s}$, intrinsics $\mathbf K_v$, and calibrated
camera-to-reconstruction transforms $\mathbf C_{v,s}\in SE(3)$.
All 3D positions and displacements are expressed in a shared metric
reconstruction frame, fixed throughout each clip and related to the robot
base by a known rigid transform.
Cameras may be fixed or follow supplied trajectories.

An articulated robot mesh is given, and joint configurations $\mathbf q_t$
and normalized gripper states $g_t\in[0,1]$ are supplied at source and query
times.
Forward kinematics (FK), followed by the fixed frame conversion, yields
link-to-reconstruction transforms $\mathbf T_\ell(t)\in SE(3)$.
Robot-mounted camera poses follow from these transforms and fixed hand--eye
calibration.

Our goal is to reconstruct a time-queryable Gaussian field
\begin{equation}
\mathcal G(t)=
\left\{
\boldsymbol\mu_i(t),
\boldsymbol\Sigma_i(t),
\alpha_i(t),
\mathbf c_i(t,\boldsymbol\omega)
\right\}_{i=1}^{N},
\label{eq:field}
\end{equation}
where the attributes denote center, covariance, opacity, and color for
viewing direction $\boldsymbol\omega$, respectively.
Gaussian identities persist across query times.
The field is rendered as
$\hat{\mathbf I}_{v,t}
=
\mathcal R(\mathcal G(t);\mathbf K_v,\mathbf C_{v,t})$,
where $\mathcal R$ denotes differentiable Gaussian rasterization~\citep{kerbl2023gaussians}.

\section{RoGSW4RLD: Robot-Aware Metric 4D Lifting}
\label{sec:method_overview}

\begin{figure}[tbp]
\centering

\includegraphics[width=1\linewidth]{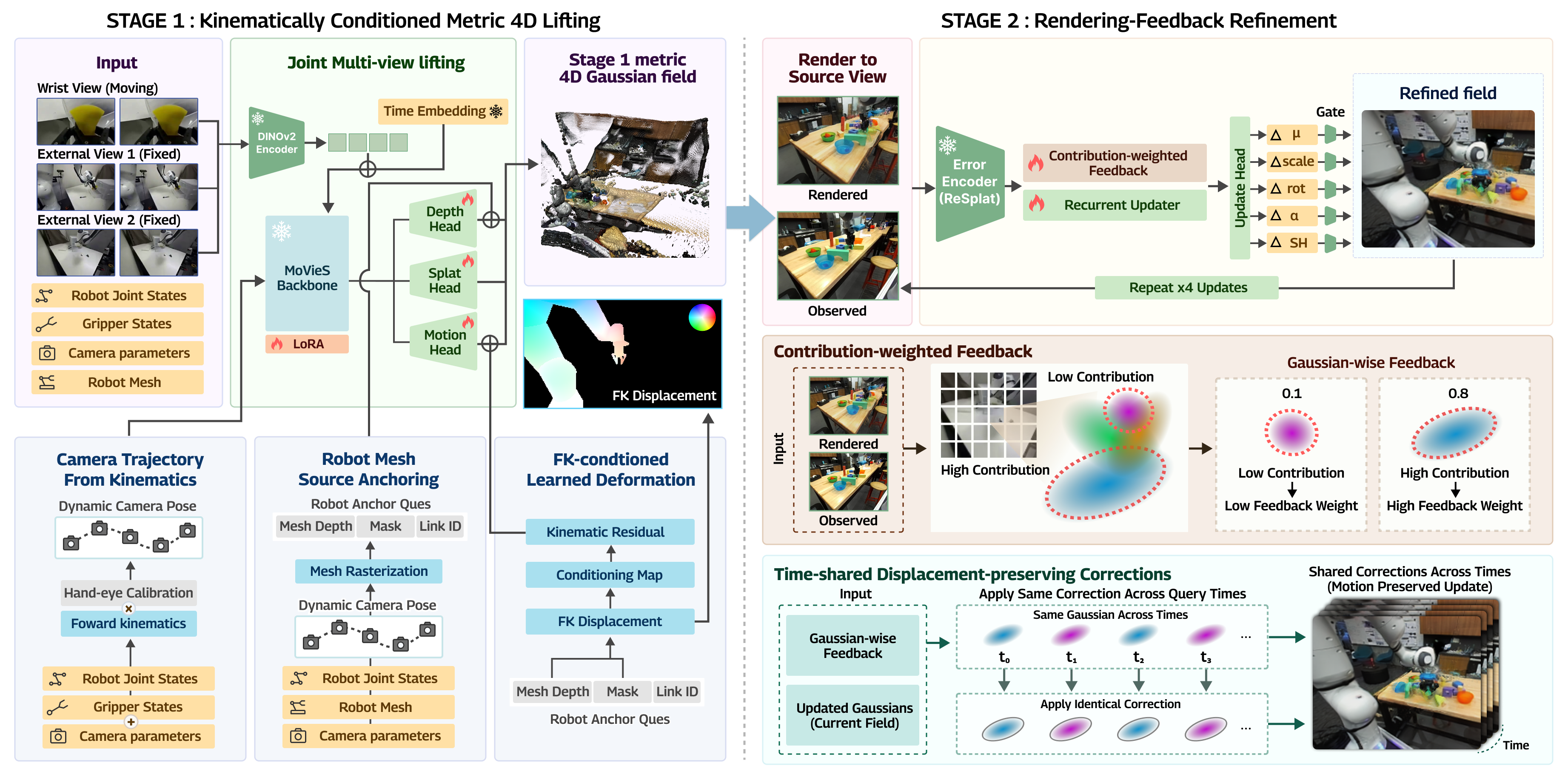}
\caption{
\textbf{Overview of RoGSW4RLD.}
Stage~1 performs kinematically conditioned metric 4D lifting by jointly processing multi-view videos with calibrated camera trajectories, robot-mesh source anchoring, and FK-conditioned learned deformation.
Stage~2 recurrently refines the resulting Gaussian field using contribution-weighted rendering feedback, while time-shared corrections preserve each Gaussian's temporal displacement across query times.
}
\label{fig:architecture}
\end{figure}

Stage~1 combines multi-camera observations with robot mesh source anchoring
and FK-conditioned deformation to form a shared metric field.
Stage~2 then refines its placement and appearance using rendering feedback
and time-shared corrections (Figure~\ref{fig:architecture}).
Both stages use fixed network weights at inference, without per-scene
optimization.

\subsection{Stage 1: Kinematically Conditioned Metric 4D Lifting}
\label{sec:metric_lifting}

\paragraph{Joint multi-camera lifting.}

We initialize the backbone from MoVieS~\citep{lin2026movies}, which builds
on VGGT~\citep{wang2025vggt}.
To adapt this backbone from monocular-video reconstruction to joint
multi-camera lifting, we apply LoRA~\citep{hu2022lora} to the QKV and output
projections in its frame-wise and global attention blocks.
Source tokens are processed jointly across views and times with camera and
time encodings.
The depth and splatter heads predict source depth and base Gaussian
attributes, respectively.
Following MoVieS, a query-time-conditioned motion head predicts 3D
displacements and time-dependent Gaussian attributes.
We use a fixed conversion between internal length units and metres across
clips, without per-scene or camera-specific rescaling
(Appendix~\ref{app:coordinates}).

\noindent
\begin{minipage}[t]{0.45\linewidth}
\vspace{0pt}

\noindent\textbf{Robot mesh source anchoring.}
Each Gaussian $i$ retains its source camera $v_i$, source time $s_i$, and
source pixel $\mathbf u_i$.
We rasterize the articulated robot mesh at each source state to obtain a
depth anchor map and per-pixel link identities.
During both training and inference, we select source depth $D_i$ from mesh
depth at valid robot-support pixels and from the depth-head prediction
elsewhere.
Back-projecting $D_i$, measured along the source camera's optical axis,
gives the source position in the shared reconstruction frame:
\begin{equation}
\mathbf p_i
=
\mathbf R_{v_i,s_i}
\left(
D_i\mathbf K_{v_i}^{-1}\bar{\mathbf u}_i
\right)
+
\mathbf t_{v_i,s_i},
\label{eq:source_position}
\end{equation}
where $\mathbf R_{v_i,s_i}$ and $\mathbf t_{v_i,s_i}$ are the rotation and
translation of $\mathbf C_{v_i,s_i}$, and $\bar{\mathbf u}_i$ is the
homogeneous source pixel.
Each mesh-anchored Gaussian also retains its link identity $\ell_i$ for the
FK-conditioned deformation described next.
Support and boundary criteria are detailed in
Appendix~\ref{app:implementation}.

\end{minipage}
\hfill
\begin{minipage}[t]{0.52\linewidth}
\vspace{0pt}
\centering

\captionsetup{type=figure}
\setcounter{subfigure}{0}

\begin{tabular}{@{}c@{}}

\begin{subfigure}[t]{\linewidth}
    \centering
    \includegraphics[width=\linewidth]
    {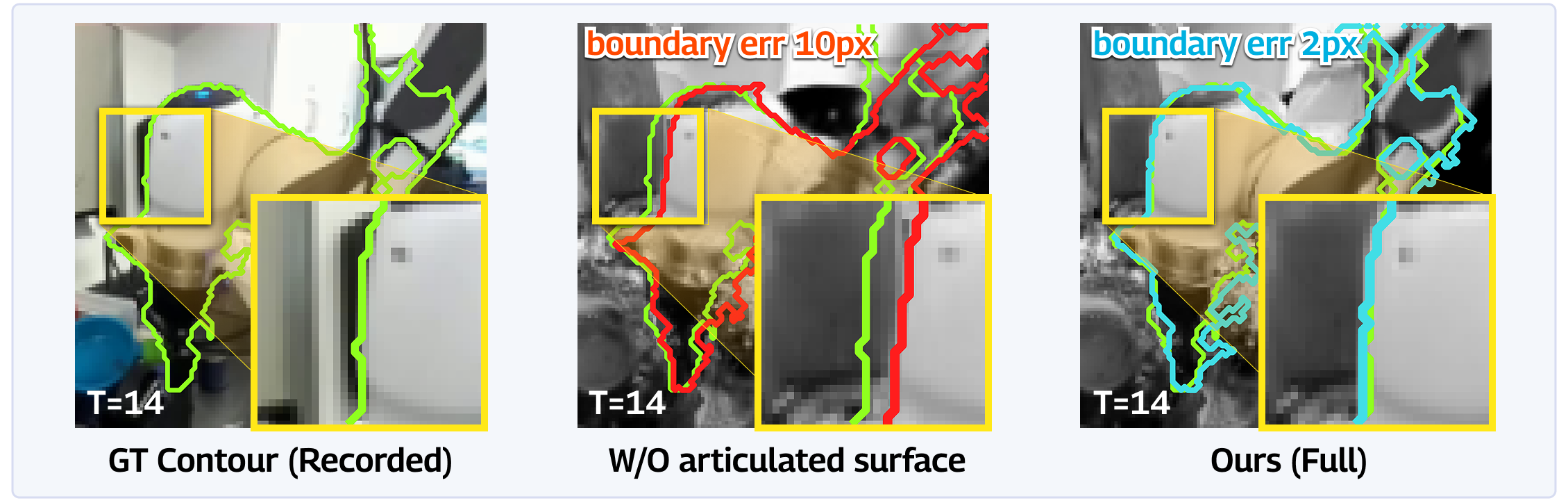}
    \caption{Robot Mesh Anchoring (Where the robot is)}
    \label{fig:source_qualitative_a}
\end{subfigure}

\\[4pt]

\begin{subfigure}[t]{\linewidth}
    \centering
    \includegraphics[width=\linewidth]
    {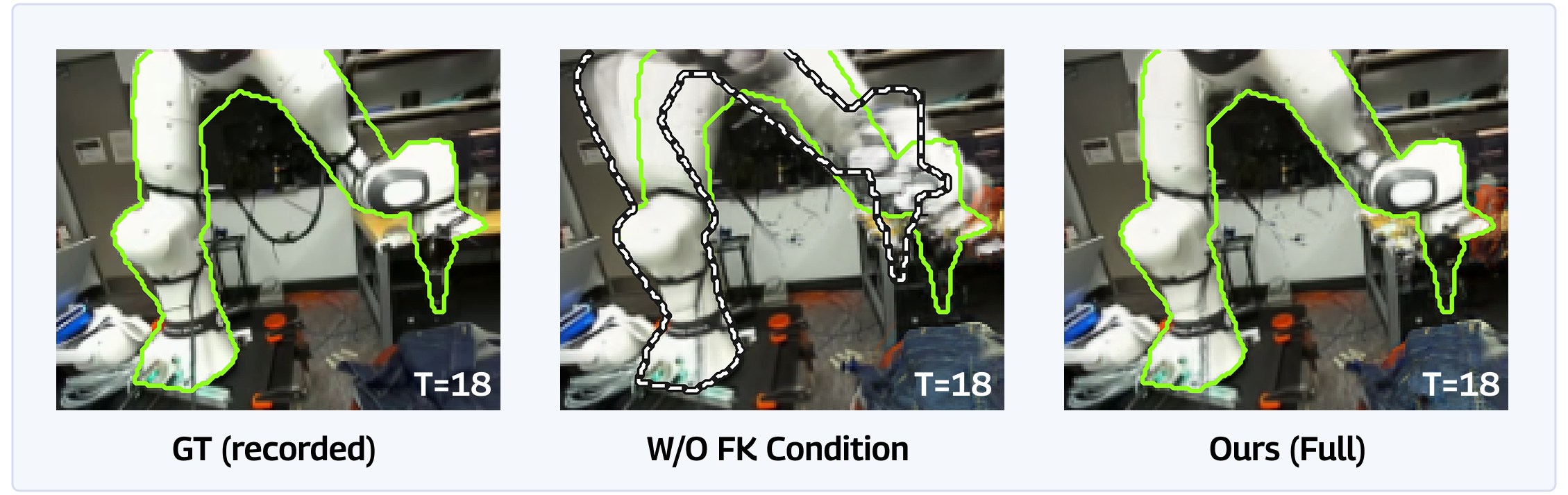}
    \caption{FK Conditioning (How the robot moves)}
    \label{fig:source_qualitative_b}
\end{subfigure}

\end{tabular}

\vspace{-2pt}

\addtocounter{figure}{-1}
\captionof{figure}{
\textbf{Robot-aware geometry and motion cues.}
(a) Robot mesh anchoring improves novel-view contour alignment (cyan), providing accurate robot placement.
(b) FK conditioning suppresses remnants of the earlier pose (dashed white) and follows the query pose (green).
}
\label{fig:source_qualitative}
\end{minipage}

\paragraph{FK-conditioned learned deformation.}

For a mesh-anchored Gaussian $i$ associated with link $\ell_i$, the FK
displacement from source time $s_i$ to query time $t$ is
\begin{equation}
\mathbf d_i^{\mathrm{FK}}(t)
=
\mathbf T_{\ell_i}(t)
\mathbf T_{\ell_i}(s_i)^{-1}
\mathbf p_i
-
\mathbf p_i,
\label{eq:fk_displacement}
\end{equation}
where rigid transforms act on 3D points through homogeneous coordinates.
We construct an FK conditioning map $\boldsymbol\kappa_{v,s}(t)$ aligned
with each source image grid.
It encodes robot support, mesh depth, and per-pixel 3D FK displacements;
auxiliary cues are detailed in Appendix~\ref{app:implementation}.

Rather than applying FK displacements directly, we use this map to augment
the motion head's visual displacement prediction and obtain the query-time
center:
\begin{align}
\mathbf d_i(t)
&=
\mathbf d_i^{\mathrm{vis}}(t)
+
\left[
\mathcal B\!\left(
\boldsymbol\kappa_{v_i,s_i}(t)
\right)
\right]_{\mathbf u_i},
\label{eq:conditioned_displacement}\\
\boldsymbol\mu_i^{(0)}(t)
&=
\mathbf p_i+\mathbf d_i(t),
\label{eq:stage1_center}
\end{align}
where $\mathbf d_i^{\mathrm{vis}}(t)$ is the visual displacement prediction
and $\mathcal B$ is a convolutional residual branch with a zero-initialized
final convolution.
The same FK displacement supervises mesh-anchored Gaussians during training.
Robot and non-robot Gaussians share the query-time deformation and
attribute-decoding pathways.
Their centers and attributes together define the shared field rendered from
each query camera.
Figure~\ref{fig:source_qualitative} illustrates the roles of mesh anchoring
and FK conditioning.

\subsection{Stage 2: Displacement-Preserving Rendering-Feedback Refinement}
\label{sec:rendering_refinement}

Stage~2 adapts the recurrent refiner of ReSplat~\citep{xu2026resplat} to
correct placement and appearance errors revealed by source-view renderings.
Stage~1 remains frozen, while source bindings, camera geometry, and the shared
reconstruction frame remain fixed.

\paragraph{Contribution-weighted rendering feedback.}

At refinement update $r$, where $r=0$ denotes the Stage~1 field, we render
the full field $\mathcal G^{(r)}(s)$ at each source time $s$ from all source
cameras.
A frozen error encoder converts discrepancies between rendered and observed
RGB images into contextual features
$\mathbf E_s^{(r)}\in\mathbb R^{P_s\times F}$, where $P_s$ counts pixels
across cameras and $F$ is the feature dimension.
To assign feedback according to each Gaussian's current visibility and
overlap with other Gaussians, we use an alpha-compositing contribution matrix
$\mathbf W_s^{(r)}\in\mathbb R^{P_s\times N}$, whose rows correspond to
pixels and columns to Gaussians.
This matrix is recomputed after each update.

For Gaussians with $s_i=s$, we compute contribution mass
$m_i^{(r)}(s)$ and pool feedback only when
$m_i^{(r)}(s)>\epsilon$:
\begin{align}
m_i^{(r)}(s)
&=
\left[
\left(\mathbf W_s^{(r)}\right)^\top
\mathbf 1
\right]_i,
\label{eq:contribution_mass}\\
\mathbf f_i^{(r)}(s)
&=
\frac{
\left[
\left(\mathbf W_s^{(r)}\right)^\top
\mathbf E_s^{(r)}
\right]_i
}{
m_i^{(r)}(s)
},
\label{eq:pooled_feedback}
\end{align}
where $\epsilon>0$ excludes negligible contributions.
The recurrent updater combines this contribution-normalized feedback with
current Gaussian attributes and its hidden state to predict corrections.

\paragraph{Time-shared displacement-preserving corrections.}

Stage~1 uses visual and FK cues to learn displacements that move Gaussians
from different source views and times to a common query time.
Independent position updates across query times could improve individual
renderings but alter these learned displacements.
To preserve them while refining Gaussian centers, we apply an accumulated
position correction $\boldsymbol\delta_i^{(r)}$ to each Gaussian after $r$
updates, shared across query times and rendering cameras:
\begin{equation}
\boldsymbol\mu_i^{(r)}(t)
=
\boldsymbol\mu_i^{(0)}(t)
+
\boldsymbol\delta_i^{(r)},
\qquad
\boldsymbol\delta_i^{(0)}=\mathbf 0.
\label{eq:shared_position_update}
\end{equation}
Thus, for any $t$ and $t'$,
\begin{equation}
\boldsymbol\mu_i^{(r)}(t)
-
\boldsymbol\mu_i^{(r)}(t')
=
\boldsymbol\mu_i^{(0)}(t)
-
\boldsymbol\mu_i^{(0)}(t').
\label{eq:displacement_preservation}
\end{equation}

The updater also predicts time-shared residuals for scale, rotation, opacity,
and spherical-harmonic coefficients.
These residuals use the corresponding Stage~1 parameterizations and are
applied to the attributes evaluated at each query time.

\subsection{Training and Inference}
\label{sec:training}

\paragraph{Training.}

During Stage~1 training, we freeze the original backbone weights and optimize
the LoRA adapters, prediction heads, and FK-conditioning branch using
rendering, metric-depth, 3D-track, robot-FK, and opacity-coverage supervision.
We then freeze Stage~1 and the observation and error encoders to train the
Stage~2 recurrent updater and residual gates.
Its objective combines rendering, metric-depth, robot-position,
rendered-motion, track-preservation, and coverage losses.
Paired-right views contribute training losses but are excluded from
reconstruction tokens and recurrent feedback.
Stereo depth and 3D track references are used for supervision and evaluation,
not as inference inputs.
Stage~2 is trained with one to four refinement updates.
Further training details are given in Appendix~\ref{app:implementation}.

\paragraph{Inference.}

At inference, we use four refinement updates with fixed network weights and
without per-scene optimization.
An optional latent-token adapter replaces only the RGB-token pathway,
leaving the remaining RGB pathways and both reconstruction stages unchanged
(Appendix~\ref{app:latent_interface}).

\section{Experiments}
\label{sec:experiments}

\subsection{Experimental Setup}
\label{sec:eval_setup}

We evaluate 256 DROID episodes from ILIAD and
PennPAL~\citep{khazatsky2024droid}, held out from training and validation of
both stages and the optional latent adapter.
Episodes are selected from the official PointWorld test
manifest~\citep{huang2026pointworld} before inspecting outputs.
Three left-eye cameras supply reconstruction inputs; paired right-eye cameras
provide held-out viewpoints.

Recorded and Cosmos~3~\citep{nvidia2026cosmos3} rollout settings share
episodes, a 17-frame interval, calibration, robot trajectories, query times,
and recorded targets.
Rollouts are conditioned on the initial three-view observation and a 16-step
FK-derived end-effector action chunk; reconstruction retains the original
joint and gripper trajectories without inverse kinematics.
RoGSW4RLD and MoVieS use frames $\{0,4,8,12,16\}$; 4DGT and SoM use all 17.
A fixed plan samples up to eight query requests per clip, evaluating duplicate
timestamps once.
Source views denote cameras, not input times.

We compare with MoVieS~\citep{lin2026movies}, 4DGT~\citep{xu20254dgt}, and
Shape of Motion (SoM)~\citep{wang2025shapeofmotion}.
Each follows its released monocular setting, reconstructing camera streams
separately and rendering the resulting fields after calibrated merging.
MoVieS and SoM receive one source-stereo scale scalar per camera;
RoGSW4RLD and 4DGT receive no test-time stereo depth.
Held-out RGB is not directly supplied, and no field is fitted to query-view
depth.
SoM completes 246 clips; others complete 256.

We report source- and novel-view PSNR, LPIPS~\citep{zhang2018lpips}, and
depth AbsRel.
Robot EPE measures displacement; robot depth error measures placement on
fixed support.
Depth references are stereo estimates, not sensor ground truth.
We average image scores over cameras and distinct times within each clip,
then equally across clips.
Confidence intervals use 10,000 clip-bootstrap resamples, reflecting
test-clip rather than training-seed variability.
Appendix~\ref{app:evaluation_protocol} details the protocol.

\begin{table}[!t]
\centering
\caption{\textbf{Quantitative results on recorded observations and
Cosmos~3 rollouts.}
RoGSW4RLD improves rendering, metric depth, and robot-displacement
accuracy over camera-wise baselines.
Both input settings are evaluated against recorded references.}
\label{tab:reconstruction_main}
\vspace{2pt}

\setlength{\tabcolsep}{2.5pt}
\renewcommand{\arraystretch}{1.08}

\resizebox{\textwidth}{!}{%
\begin{tabular}{
@{}l
ccc
@{\hspace{10pt}}
ccc
@{\hspace{10pt}}
c
@{\hspace{10pt}}
cc@{}
}

\noalign{\hrule height 1.2pt}

\raisebox{-1.35ex}{\textbf{Method}}
&
\multicolumn{3}{c}{
    \shortstack{
        \rule{0pt}{2.2ex}\textbf{Source view}\\[0.25ex]
        \rule{1.40in}{0.4pt}
    }
}
&
\multicolumn{3}{c}{
    \shortstack{
        \rule{0pt}{2.2ex}\textbf{Novel view}\\[0.25ex]
        \rule{1.40in}{0.4pt}
    }
}
&
\shortstack{
    \rule{0pt}{2.2ex}\textbf{Robot}\\[0.25ex]
    \vphantom{\rule{0.45in}{0.4pt}}
}
&
\multicolumn{2}{c}{
    \shortstack{
        \rule{0pt}{2.2ex}\textbf{Efficiency}\\[0.25ex]
        \rule{0.88in}{0.4pt}
    }
}
\\[0.4ex]

&
PSNR $\uparrow$
&
LPIPS $\downarrow$
&
AbsRel $\downarrow$
&
PSNR $\uparrow$
&
LPIPS $\downarrow$
&
AbsRel $\downarrow$
&
EPE {\footnotesize (mm)} $\downarrow$
&
Time {\footnotesize (s)} $\downarrow$
&
VRAM {\footnotesize (GiB)} $\downarrow$
\\[1.2ex]

\hline

\multicolumn{10}{@{}l}{\textit{Recorded observations}} \\[1pt]

MoVieS$^\dagger$
& 19.71 & 0.306 & 0.306
& 13.93 & 0.489 & 0.373
& 22.2 & 1.04 & 4.46 \\

4DGT
& 16.54 & 0.456 & 0.600
& 12.90 & 0.584 & 0.681
& 31.0 & 4.80 & 3.89 \\

SoM$^\dagger$
& 14.22 & 0.443 & 0.383
& 12.14 & 0.557 & 0.429
& 32.2 & 1713 & 2.35 \\

Ours (Stage~1)
& 21.92 & 0.171 & 0.160
& 15.52 & \textbf{0.363} & 0.200
& \textbf{8.6} & 0.59 & 5.20 \\

Ours (Stage~1+2)
& \textbf{25.46} & \textbf{0.166} & \textbf{0.159}
& \textbf{16.08} & 0.371 & \textbf{0.197}
& \textbf{8.6} & 9.08 & 13.11 \\

\hline

\multicolumn{10}{@{}l}{\textit{Cosmos~3 rollouts}} \\[1pt]

MoVieS$^\dagger$
& 18.33 & 0.348 & 0.343
& 13.69 & 0.503 & 0.404
& 22.6 & 1.04 & 4.46 \\

4DGT
& 15.78 & 0.483 & 0.625
& 12.78 & 0.591 & 0.697
& 32.3 & 4.80 & 3.89 \\

SoM$^\dagger$
& 13.92 & 0.463 & 0.391
& 12.05 & 0.564 & 0.440
& 37.6 & 1985 & 2.39 \\

Ours (Stage~1)
& 19.87 & 0.235 & 0.182
& 15.04 & \textbf{0.394} & 0.218
& \textbf{8.7} & 0.59 & 5.20 \\

Ours (Stage~1+2)
& \textbf{22.12} & \textbf{0.231} & \textbf{0.181}
& \textbf{15.52} & 0.402 & \textbf{0.215}
& \textbf{8.7} & 9.08 & 13.11 \\

\noalign{\hrule height 1.2pt}

\end{tabular}%
}

\par\smallskip
{\fontsize{8}{9.5}\selectfont\raggedright
$n=256$ (246 successful clips for SoM).
$\dagger$: one scale factor per camera from recorded source stereo depth. \\
See Appendix~\ref{app:evaluation_details} for evaluation details
and Appendix~\ref{app:efficiency} for method-specific timing
and memory measurements.
\par}

\end{table}

\subsection{Reconstruction Results and Efficiency}
\label{sec:eval_results}
\label{sec:eval_refinement}
\label{sec:eval_rollout}

Table~\ref{tab:reconstruction_main} shows improvements over camera-wise
reconstruction followed by calibrated merging.
Stage~1 and the full model both outperform the baselines in rendering,
metric depth, and robot displacement.
On recorded inputs, our final model improves novel-view PSNR over MoVieS by
2.15~dB, reduces AbsRel from 0.373 to 0.197, and lowers robot EPE from
22.2 to 8.6~mm.
Most geometric and motion gains are already present in Stage~1.
These are system-level comparisons, since architectures, training, and robot
inputs differ.

\begin{figure}[!b]
    \centering
    \includegraphics[width=\linewidth]{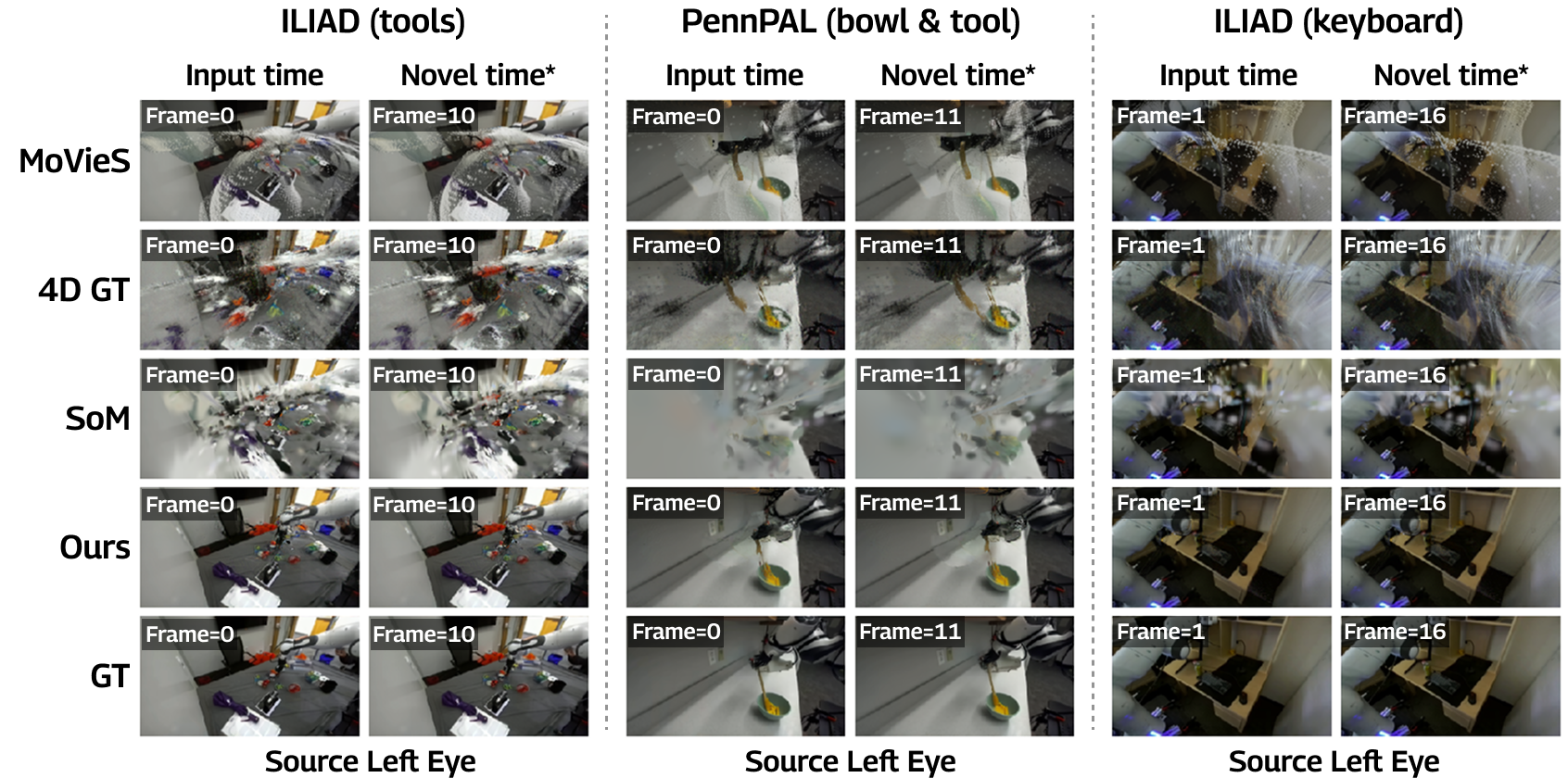}
\caption{
\textbf{Qualitative comparison on action-conditioned Cosmos~3 rollouts.}
Across ILIAD and PennPAL scenes, RoGSW4RLD produces cleaner reconstructions with less ghosting and sharper robot and object boundaries than the baselines at both input and novel times.
The bottom row shows the corresponding generated Cosmos~3 frames, rather than recorded observations; asterisks denote non-input query times for RoGSW4RLD and MoVieS.
}
    \label{fig:cosmos_qualitative}
    \end{figure}

With generated rollouts, final novel-view PSNR decreases by 0.56~dB and
AbsRel increases by 0.018 relative to recorded inputs.
Nevertheless, both remain better than MoVieS:
15.52 versus 13.69~dB and 0.215 versus 0.404.
The reconstruction advantage therefore persists when the visual future is
generated.
Evaluation against recorded futures combines rollout and reconstruction
errors.
Robot trajectories are unchanged, so stable robot EPE does not establish
accurate visual motion generation.
Figure~\ref{fig:cosmos_qualitative} illustrates reduced ghosting in our
source-view reconstructions of generated rollouts; recorded-input novel-view
comparisons appear in Appendix~\ref{app:recorded_qualitative}.

Stage~2 improves source and novel-view PSNR by 3.54 and 0.56~dB on recorded
inputs, respectively, and by 2.25 and 0.48~dB on rollouts.
In both settings, novel-view scores show a 0.003 decrease in AbsRel but worse
LPIPS, while robot EPE remains nearly unchanged.
Refinement thus primarily improves photometric fitting.
Time-shared corrections preserve Gaussian center displacements, but rendered
motion can change with compositing weights.

Stage~1 requires 0.59~s and 5.20~GiB per clip, versus 9.08~s and
13.11~GiB with four refinement updates.
Both use fixed inference weights without per-scene optimization.
A 32-clip recurrence diagnostic shows diminishing PSNR gains beyond two
updates (Appendix~\ref{app:recurrence});
Appendix~\ref{app:efficiency} details timing and memory.

\subsection{Ablation Study}
\label{sec:eval_ablation}
\label{sec:eval_joint}
\label{sec:eval_kinematics}

We examine cross-camera interaction, rendering-feedback refinement, robot mesh source
anchoring, and FK conditioning through frozen-checkpoint interventions
(Table~\ref{tab:controlled_ablation}).
These diagnose inference-time dependence rather than compare separately
retrained variants.

\begin{table}[!t]
\centering
\caption{\textbf{Frozen-checkpoint component interventions.}
(a) Joint lifting improves novel-view geometry; refinement mainly
improves RGB fitting.
(b) Mesh anchoring improves robot placement, while FK conditioning
improves displacement accuracy.}
\vspace{2pt}
\label{tab:controlled_ablation}

\scriptsize
\renewcommand{\arraystretch}{1.07}

\begin{minipage}[t]{0.48\linewidth}
\centering
{\normalsize (a) Joint multi-camera lifting and refinement}
\setlength{\tabcolsep}{0.6pt}

\begin{tabular}
{@{}ccccccc@{}}
\noalign{\hrule height 0.9pt}
\shortstack{Cross\\camera}
& Stage2
& \shortstack{Source\\PSNR $\uparrow$}
& \shortstack{Novel\\PSNR $\uparrow$}
& \shortstack{Novel\\AbsRel $\downarrow$}
& \shortstack{Robot EPE\\(mm) $\downarrow$}
& \shortstack{Robot depth\\err. (mm) $\downarrow$} \\
\hline

\multicolumn{7}{c}{Recorded Observations} \\
\hline

$\checkmark$ & $\checkmark$
& \textbf{25.46} & \textbf{16.08} & \textbf{0.197}
& \textbf{8.6} & 13.7 \\

$\checkmark$ & \textemdash
& 21.92 & 15.52 & 0.200
& \textbf{8.6} & \textbf{13.4} \\

\textemdash & $\checkmark$
& 23.71 & 12.93 & 0.392
& 8.8 & 101.6 \\

\textemdash & \textemdash
& 19.64 & 12.64 & 0.391
& 8.8 & 100.6 \\

\hline
\multicolumn{7}{c}{Cosmos~3 Rollouts} \\
\hline

$\checkmark$ & $\checkmark$
& \textbf{22.12} & \textbf{15.52} & \textbf{0.215}
& \textbf{8.72} & 15.5 \\

$\checkmark$ & \textemdash
& 19.87 & 15.04 & 0.218
& \textbf{8.72} & \textbf{15.2} \\

\textemdash & $\checkmark$
& 21.37 & 12.75 & 0.391
& 8.90 & 97.9 \\

\textemdash & \textemdash
& 18.38 & 12.50 & 0.390
& 8.92 & 97.2 \\

\noalign{\hrule height 0.9pt}
\end{tabular}
\end{minipage}
\hfill
%
\begin{minipage}[t]{0.48\linewidth}
\centering
{\normalsize (b) Robot geometry and motion}
\setlength{\tabcolsep}{0.6pt}

\begin{tabular}
{@{}ccccccc@{}}
\noalign{\hrule height 0.9pt}
\shortstack{Mesh\\anchoring}
& \shortstack{FK\\cond.}
& \shortstack{Source\\PSNR $\uparrow$}
& \shortstack{Novel\\PSNR $\uparrow$}
& \shortstack{Novel\\AbsRel $\downarrow$}
& \shortstack{Robot EPE\\(mm) $\downarrow$}
& \shortstack{Robot depth\\err. (mm) $\downarrow$} \\
\hline

\multicolumn{7}{c}{Recorded Observations} \\
\hline

$\checkmark$ & $\checkmark$
& \textbf{25.46} & \textbf{16.08} & 0.197
& \textbf{8.64} & \textbf{13.7} \\

$\checkmark$ & \textemdash
& 23.26 & 15.77 & 0.206
& 30.15 & 22.1 \\

\textemdash & $\checkmark$
& 25.29 & 16.07 & \textbf{0.196}
& 8.81 & 37.6 \\

\textemdash & \textemdash
& 23.29 & 15.79 & 0.207
& 30.12 & 42.9 \\

\hline
\multicolumn{7}{c}{Cosmos~3 Rollouts} \\
\hline

$\checkmark$ & $\checkmark$
& \textbf{22.12} & 15.52 & \textbf{0.215}
& \textbf{8.72} & \textbf{15.5} \\

$\checkmark$ & \textemdash
& 21.20 & 15.34 & 0.223
& 30.16 & 24.5 \\

\textemdash & $\checkmark$
& 22.02 & \textbf{15.56} & 0.218
& 8.90 & 56.6 \\

\textemdash & \textemdash
& 21.21 & 15.39 & 0.227
& 30.14 & 62.7 \\

\noalign{\hrule height 0.9pt}
\end{tabular}
\end{minipage}

\par\smallskip
{\scriptsize\raggedright
All interventions are applied at inference without retraining
($n=256$);
(b) uses Stage~1+2. \\ 
Robot depth error is source-view depth MAE on fixed robot support. \\
See Appendix~\ref{app:evaluation_details} for evaluation details and Appendix~\ref{app:intervention_details} for intervention definitions
and Stage~1 results.
\par}

\end{table}


\paragraph{Joint lifting and refinement.}

We block cross-camera interaction while retaining within-camera temporal
interaction, calibration, and robot inputs, then merge the resulting
camera-wise fields.
In Table~\ref{tab:controlled_ablation}(a), camera-wise refinement raises
recorded source PSNR from 19.64 to 23.71~dB but leaves novel-view AbsRel at
0.391--0.392.
Joint Stage~1 instead achieves 0.200 AbsRel at 21.92~dB source PSNR.
Thus, stronger source-view fitting does not recover the held-out geometry
obtained through joint multi-camera lifting.

Rollouts show the same separation.
Stage~2 raises source PSNR of the camera-wise reconstruction from 18.38 to
21.37~dB, while novel-view AbsRel remains at 0.390--0.391.
Joint Stage~1 instead reaches 0.218 AbsRel.
Because this intervention blocks cross-camera interaction only at inference,
it should not be interpreted as training three independent monocular models
from scratch.

\paragraph{Robot geometry and kinematics.}

We disable mesh anchoring, FK conditioning, or both from the same
frozen checkpoint.
When mesh anchoring is disabled, supported robot geometry falls back to the
learned visual geometry path while FK conditioning remains available.
Conversely, removing FK conditioning retains mesh anchoring.
Camera calibration is unchanged.
FK-conditioning removal is supported by conditioning dropout during training,
whereas mesh anchoring removal and removal of both are diagnostic
out-of-distribution interventions.

Table~\ref{tab:controlled_ablation}(b) distinguishes placement from
displacement.
On recorded inputs, removing FK conditioning raises EPE from 8.64 to
30.15~mm.
Removing mesh anchoring instead raises robot depth error from 13.7 to
37.6~mm, while EPE remains at 8.81~mm.
Removing both degrades both metrics.

Rollouts show the same pattern.
FK removal raises EPE from 8.72 to 30.16~mm.
Removing mesh anchoring raises robot depth error from 15.5 to 56.6~mm while
EPE remains at 8.90~mm.
The larger rollout placement penalty is observed, but its cause is not
isolated by these interventions.
Scene-wide AbsRel does not uniformly improve with mesh anchoring because the
intervention targets the robot region.

Figure~\ref{fig:source_qualitative} visualizes the same distinction at
non-input query times.
Without mesh anchoring, the metric robot surface shows a larger disparity
shift.
Without FK conditioning, stale robot content remains near an earlier pose.
Appendix~\ref{app:joint_ablation} reports the stage-wise intervention
details.
Analytic FK transport slightly improves robot EPE, so learned deformation is
not claimed to be more accurate than analytic kinematics
(Appendix~\ref{app:rigid_policy}).
Its role is instead to keep robot and non-robot dynamics within a shared
learned deformation representation.

\section{Conclusion}
\label{sec:conclusion}

We presented RoGSW4RLD, a feed-forward framework that converts synchronized robot-camera rollouts into a shared, time-queryable metric 4D Gaussian field. Rather than learning a separate geometric transition model, it directly reconstructs visual futures from existing video world models. Our core novelty lies in jointly forming this metric field across moving wrist and fixed external cameras by integrating articulated robot geometry and kinematics. This approach circumvents the cross-view inconsistencies of post-hoc alignment, substantially enhancing geometric consistency and motion accuracy. Additionally, our displacement-preserving refinement boosts visual quality without altering the estimated temporal dynamics.

Evaluations on recorded DROID observations and action-conditioned Cosmos~3 rollouts validate these advantages across novel-view synthesis, metric geometry, and displacement accuracy. These results show that predicted multi-camera videos can be converted into coherent, queryable metric scenes without per-scene optimization or retraining the underlying world model. Ultimately, RoGSW4RLD bridges video predictions and explicit spatial representations to facilitate downstream robotic reasoning. Future work will extend this framework beyond calibrated setups to unconstrained sensing, longer horizons, and diverse robotic configurations.



\subsection*{AI Use Statement}
In this work, we used generative AI tools to support research ideation and discussions of experimental design, identify relevant literature, and draft and polish portions of the manuscript.

We have not used generative AI tools for synthetic dataset generation, mathematical proofs, or any purposes beyond those described above.

At least two authors reviewed all AI-assisted content and revised it where necessary.
This review included checking references against the original sources and verifying that technical descriptions and interpretations were consistent with our methods and experimental results.
The authors retain full responsibility for the accuracy and integrity
of the paper and its associated artifacts.

\subsection*{Reproducibility Statement}

Section~\ref{sec:method_overview} and Appendix~\ref{app:implementation}
describe the architecture, losses, and training schedules of both stages,
while Appendix~\ref{app:coordinates} specifies the coordinate and unit
conventions.
Section~\ref{sec:eval_setup} and Appendix~\ref{app:evaluation_protocol}
document the evaluation cohort, input protocol, baseline adaptations, and
metrics, and Table~\ref{tab:main_uncertainty} reports confidence intervals.
Code will be released publicly at a later date.
Recorded data and evaluation references are derived from the public DROID
and PointWorld releases; generated rollouts use public releases of Cosmos~3
and Wan2.2.


\bibliography{iclr2027_conference}
\bibliographystyle{iclr2027_conference}

\clearpage

\appendix
\section*{Appendix Contents}

\begingroup
\small
\setlength{\parskip}{0pt}
\newcommand{\appcontentssection}[2]{%
  \par\vspace{0.45em}\noindent
  \hyperref[#1]{\textbf{\ref*{#1}\quad #2}}%
  \dotfill\hyperref[#1]{\pageref*{#1}}\par
}
\newcommand{\appcontentssubsection}[2]{%
  \noindent\hspace*{1.5em}%
  \hyperref[#1]{\ref*{#1}\quad #2}%
  \dotfill\hyperref[#1]{\pageref*{#1}}\par
}

\appcontentssection{app:implementation}{Implementation Details}
\appcontentssubsection{app:coordinates}{Coordinate system and robot geometry}
\appcontentssubsection{app:stage1_training}{Stage 1 architecture and training}
\appcontentssubsection{app:stage2_training}{Stage 2 architecture and training}

\appcontentssection{app:evaluation_details}{Evaluation Protocol}
\appcontentssubsection{app:cohort_selection}{Evaluation cohort}
\appcontentssubsection{app:temporal_sampling}{Cameras and temporal sampling}
\appcontentssubsection{app:inference_inputs}{Inference inputs and evaluation references}
\appcontentssubsection{app:baseline_protocols}{Baseline adaptations and coverage}
\appcontentssubsection{app:metric_definitions}{Metric definitions}

\appcontentssection{app:extended_results}{Additional Baseline Comparisons and Uncertainty}
\appcontentssubsection{app:recorded_qualitative}{Qualitative reconstruction from recorded observations}
\appcontentssubsection{app:additional_metrics}{Metrics omitted from the main table}
\appcontentssubsection{app:confidence_intervals}{Confidence intervals}
\appcontentssubsection{app:frame_count}{Number of input frames}
\appcontentssubsection{app:scale_diagnostic}{Scale-adjusted depth diagnostic}

\appcontentssection{app:ablations}{Additional Design Analyses}
\appcontentssubsection{app:joint_ablation}{Intervention definitions and Stage 1 effects}
\appcontentssubsection{app:rigid_policy}{Learned motion and analytic FK transport}
\appcontentssubsection{app:recurrence}{Number of rendering-feedback refinement updates}
\appcontentssubsection{app:efficiency}{Reconstruction and rendering cost}

\appcontentssection{app:latent_interface}{Optional Latent-Token Interface}
\appcontentssubsection{app:adapter_training}{Adapter and training}
\appcontentssubsection{app:rgb_token_effect}{Effect of replacing RGB tokens}
\appcontentssubsection{app:spatial-resolution-diagnostics}{Spatial-resolution diagnostic}

\appcontentssection{app:limitations}{Scope and Limitations}
\endgroup

\clearpage

\section{Implementation Details}
\label{app:implementation}

\subsection{Coordinate system and robot geometry}
\label{app:coordinates}

\paragraph{Coordinate convention.}
Supplied camera calibrations and FK define camera and link poses in the robot
base frame.
For each clip, we set the reconstruction frame's origin and orientation to
those of the wrist camera at the first source time $s_0=\min\mathcal S$.
This frame remains fixed throughout the clip, even as the wrist camera moves.
Let $\mathbf C^{\mathrm{base}}_{v,t}$ and
$\mathbf T^{\mathrm{base}}_\ell(t)$ denote camera-to-base and link-to-base
transforms.
The fixed base-to-reconstruction transform is
$\mathbf H=(\mathbf C^{\mathrm{base}}_{\mathrm{wrist},s_0})^{-1}$, giving
\begin{equation}
\mathbf C_{v,t}=\mathbf H\mathbf C^{\mathrm{base}}_{v,t},
\qquad
\mathbf T_\ell(t)=\mathbf H\mathbf T^{\mathrm{base}}_\ell(t).
\label{eq:coordinate_convention}
\end{equation}
The same frame is used for all source views, query times, Gaussian centers,
and 3D displacements.
Wrist-camera poses follow the supplied joint states and fixed hand--eye
calibration.
Joint configurations $\mathbf q_t$ and normalized gripper states
$g_t\in[0,1]$ are supplied at source and query times, not predicted by the
lifter.

\paragraph{Internal units.}
The coordinate definitions above use metres.
Network and rasterizer computations use a fixed conversion of 3.5 internal
units per metre across clips.
Geometric positions, displacements, Gaussian scales, and camera/link
translations use this conversion consistently; rotation matrices are not
rescaled.
This changes numerical units, not the physical scale of a scene, and no
clip-specific scale normalization is applied.
Metric quantities can be expressed in the robot base frame using the inverse
fixed transform after undoing the unit conversion.

\paragraph{Robot mesh source anchoring.}
We rasterize the Panda collision mesh and Robotiq visual mesh at each source
state to obtain source-camera depth anchor maps and link identities.
For Gaussian $i$, source depth $D_i$ is selected from mesh depth at valid
robot-support pixels and from the depth-head prediction elsewhere.
This deterministic selection is used during training and inference, rather
than learned as a fusion rule.
Depth denotes source-camera optical-axis depth, not Euclidean distance to
the camera center.
We retain the resulting source position $\mathbf p_i$, mesh material-point
correspondence, link identity $\ell_i$, and fixed source binding
$(v_i,s_i,\mathbf u_i)$.
This \emph{robot mesh source anchoring} constrains supported robot geometry;
all remaining pixels use learned depth.
Cached stereo surfaces are supervision or evaluation references, not
reconstruction inputs.

\subsection{Stage 1 architecture and training}
\label{app:stage1_training}

Stage~1 is initialized from released MoVieS~\citep{lin2026movies} weights,
whose backbone builds on VGGT~\citep{wang2025vggt}.
We freeze the original backbone parameters and train LoRA~\citep{hu2022lora}
adapters, the depth, motion, and Gaussian-attribute heads, and the
FK-conditioning branch.
LoRA uses rank 32 and $\alpha_{\mathrm{LoRA}}=16$ and is applied to the QKV
and output projections in the frame-wise and global attention blocks.
Tokens from source views and times are processed jointly before geometry and
motion decoding.

The conditioning map $\boldsymbol\kappa_{v,s}(t)$ is aligned with the source
image grid and contains robot support, source mesh depth, source-to-query
3D FK displacement $\mathbf d_i^{\mathrm{FK}}(t)$, an availability flag,
and a coarse gripper-neighborhood prior.
The three displacement channels store vectors in the shared reconstruction
frame and internal length units, not image-plane flow; the source mesh-depth
channel remains in metres.
The final convolution of $\mathcal B$ is zero-initialized, initially preserving
the visual displacement prediction.
Training includes FK-conditioning dropout.
Mesh anchoring constrains source placement and Gaussian footprint, whereas
query-time centers $\boldsymbol\mu_i^{(0)}(t)=\mathbf p_i+\mathbf d_i(t)$
remain learned predictions rather than analytic FK replacements.

The Gaussian-attribute head receives backbone features and a resized RGB
shortcut; the depth head uses tokens only.
Following MoVieS, the main text uses \emph{motion head} collectively for
query-time displacement and attribute prediction.
The implementation separates these into \texttt{motion\_head}, which predicts
3D center displacement, and \texttt{motion\_splat\_head}, which predicts
query-time color, scale, rotation, and opacity from backbone features, base
Gaussian attributes, and the query-time encoding.
FK conditioning directly augments only the displacement prediction through
$\mathcal B$.
For both robot and non-robot Gaussians, query-time attributes replace static
values under the released MoVieS motion gates.
Color, scale, and rotation use a center-motion threshold of 0.005 internal
units (approximately 1.4~mm); opacity retains its separate gate.
This permits time-dependent appearance.

Training uses AdamW with $\beta=(0.9,0.95)$ and weight decay 0.05.
Peak learning rates are $4\times10^{-5}$ for backbone adapters and standard
heads, and $4\times10^{-4}$ for Gaussian-attribute and kinematic branches.
A 150-update linear warmup precedes cosine decay; gradients are clipped
at 1.0.
We use seed 73 and the final checkpoint at update 3,462, after approximately
ten data passes on four B200 GPUs.
Each clip samples two to five source times and eight query requests.
Batch size adapts to source-frame count within a budget of 80 source images
per GPU.
Neural modules use bfloat16 mixed precision; geometric operations use float32.

Supervision combines source and paired-right RGB rendering, metric depth,
3D tracks, robot-FK consistency, and opacity coverage.
Paired-right images contribute training losses, not reconstruction tokens.
Table~\ref{tab:stage1_loss_weights} lists the separately weighted loss terms.

\begin{table}[!ht]
\centering
\caption{Stage~1 loss weights. Keys identify separately weighted terms in the training configuration. Values are reported to at most six significant figures.}
\label{tab:stage1_loss_weights}
\small
\setlength{\tabcolsep}{8pt}
\renewcommand{\arraystretch}{1.08}
\begin{tabular}{@{}lrlr@{}}
\hline
Loss key & Weight & Loss key & Weight \\
\hline
\texttt{movies\_mse} & 1.0 & \texttt{movies\_lpips} & 0.5 \\
\texttt{depth\_head} & 1.0 & \texttt{track\_head} & 10.0 \\
\texttt{stereo} & 1.04051 & \texttt{track\_final} & 0.105593 \\
\texttt{fk\_head} & 0.464340 & \texttt{fk\_final} & 0.157137 \\
\texttt{metric\_final} & 0.00182939 & \texttt{metric\_right} & 0.00484463 \\
\texttt{coverage} & 0.0589678 & \texttt{coverage\_right} & 0.0648524 \\
\hline
\end{tabular}
\end{table}

\subsection{Stage 2 architecture and training}
\label{app:stage2_training}

Stage~2 freezes Stage~1 and initializes the refiner from released
ReSplat~\citep{xu2026resplat} weights.
The observation and error encoders remain frozen.
We train the recurrent updater and residual gates for 1,928 optimization
updates over two exposure epochs, sampling one to four refinement updates
during training and using four at inference.
AdamW learning rates are $10^{-4}$ for the updater and $1.521\times10^{-4}$
for the gates, with weight decay 0.01 and the released 1\% warmup schedule.

At source time $s$, the full current field is rendered across source cameras.
Feedback is assigned only to Gaussians with $s_i=s$ and contribution mass
$m_i^{(r)}(s)>\epsilon$.
Contextual errors are pooled using current alpha-compositing contributions
and normalized by contribution mass; weights are recomputed after each update.

For each eligible Gaussian, the updater predicts a 59-channel, time-shared
residual for position, scale, rotation, opacity, and spherical-harmonic
coefficients.
Shared corrections use the corresponding attribute parameterizations and
apply to Stage~1's time-specific values.
The accumulated position correction $\boldsymbol\delta_i^{(r)}$ is shared
across query times and cameras and bounded by 5~cm.
This preserves Gaussian center displacements exactly, but not rendered
motion, which can change with compositing weights.
No separate trajectory-correction branch is used.

The objective combines source and paired-right rendering with metric-depth
and absolute FK-referenced robot-position losses, plus rendered-motion,
track-preservation, and coverage terms.
These losses constrain placement and rendering, while time sharing preserves
Stage~1 center displacements.
Only source RGB enters recurrent feedback; paired-right images serve as
training loss targets, not feedback inputs.
Inference uses fixed network weights without per-scene optimization.


\section{Evaluation Protocol}
\label{app:evaluation_details}
\label{app:evaluation_protocol}

\subsection{Evaluation cohort}
\label{app:cohort_selection}

\paragraph{Lab holdout and test membership.}
Our internal split reserves all 1,026 ILIAD and PennPAL episodes for evaluation.
Neither lab contributes episodes to our training or validation of Stage~1,
Stage~2, or the optional latent adapter.
The prepared cache contains one 33-frame clip with stereo references for each
of 804 episodes.
Episodes appearing in the preparation script's training or validation cohorts
are also excluded.

We use the released DROID test assignments from
PointWorld~\citep{huang2026pointworld}, rather than regenerate the split.
The manifest \texttt{droid\_paper\_split\_manifest.json} specifies the DROID
domain, seed 42, and test fraction 0.1.
Removing the final temporal-clip suffix from each manifest key gives its
episode ID.
Membership is therefore checked at the episode level, without requiring our
evaluation interval to match the manifest's temporal clip.

\paragraph{Eligibility and sampling.}
Of the 804 cached episodes, 306 are absent from the manifest-derived test
episode set, leaving 498 eligible episodes.
We additionally require at least 24 cached frames, available camera-calibration
metadata, and finite normalized gripper values in $[0,1]$, allowing a boundary
tolerance of $10^{-6}$.
These checks exclude no further episodes.

Using \texttt{numpy.random.default\_rng(42)}, we permute eligible records
sorted by cached clip name.
We select the first 256 indices without replacement, then sort those indices
only to fix export order.
Sampling is uniform over eligible episodes, not stratified by lab or scene.
The cohort contains 129 PennPAL and 127 ILIAD episodes across 69 scenes, with
one fixed interval per episode shared by recorded and rollout evaluations.
Table~\ref{tab:cohort_selection} summarizes the selection.

\begin{table}[!ht]
\centering
\caption{Evaluation-cohort construction. Counts refer to episodes, not camera streams or temporal queries.}
\label{tab:cohort_selection}
\small
\setlength{\tabcolsep}{8pt}
\renewcommand{\arraystretch}{1.08}
\begin{tabular}{@{}lr@{}}
\hline
Selection stage & Episodes \\
\hline
ILIAD and PennPAL episodes reserved in the internal split & 1,026 \\
Episodes with a prepared clip and stereo references & 804 \\
Eligible episodes represented in the PointWorld test manifest & 498 \\
Fixed-seed evaluation sample & 256 \\
\hline
\end{tabular}
\end{table}

\paragraph{Selection record.}
The cohort is fixed before inspecting model outputs.
Before constructing rollout inputs, the preparation script writes
\texttt{selection.json}, recording the seed, candidate and eligible counts,
exclusion reasons, and selected episode IDs.
Subsequent method-specific execution failures affect evaluation coverage,
not the initial 256-episode cohort.

\subsection{Cameras and temporal sampling}
\label{app:temporal_sampling}

Each episode provides a 33-frame clip at 15~Hz.
Stored wrist and external images have resolutions $352\times640$ and
$176\times320$, respectively; all six cameras are evaluated at $154\times280$.
The wrist and two external left-eye streams provide reconstruction observations.
Their paired right-eye cameras are the held-out novel views.

Both settings use cached frames 7--23, a 17-frame interval matching one
Cosmos~3--Nano~\citep{nvidia2026cosmos3} rollout.
Frame 7 supplies the initial three-view observation, followed by 16 frames
conditioned on a 16-step FK-derived end-effector action chunk.
Reconstruction retains the supplied joint and gripper trajectories without
inverse kinematics.
Our five designated source times $\mathcal S$ have relative indices
$\{0,4,8,12,16\}$, corresponding to cached frames $\{7,11,15,19,23\}$.
RoGSW4RLD and MoVieS use these times; the main 4DGT and SoM baselines use all
17 frames per camera.
Appendix~\ref{app:frame_count} also evaluates five-frame 4DGT.

A fixed per-clip random plan samples up to eight query requests within this
interval.
With probability 0.25, it includes all five times in $\mathcal S$ and samples
the remaining requests from the interval; otherwise, all requests are sampled
directly.
Duplicate timestamps are evaluated once.
Recorded and rollout settings share query times, evaluation cameras, robot
trajectories, and recorded RGB/depth/track references.
Only the reconstruction observations change.
An input time denotes temporal inclusion in a method's input sequence;
a source view denotes camera identity, irrespective of whether the queried
time is an input time.

\subsection{Inference inputs and evaluation references}
\label{app:inference_inputs}

\begin{table}[!ht]
\centering
\caption{Signals used in the main evaluation. ``Scale only'' denotes one scalar per camera computed at the five designated source times, not dense stereo supervision during test-time reconstruction. Right-eye maps and reference tracks are not supplied directly to reconstruction.}
\label{tab:input_contract}
\footnotesize
\setlength{\tabcolsep}{3.5pt}
\renewcommand{\arraystretch}{1.12}
\begin{tabular}{@{}lcccc@{}}
\hline
Signal & Ours & MoVieS & 4DGT & SoM \\
\hline
Left-eye visual input & 5 times $\times$ 3 views & 5 / camera & 17 / camera & 17 / camera \\
Calibrated cameras & Yes & Yes & Yes & Yes \\
Joint/gripper states$^\ddagger$ & Anchoring/FK & No direct input & No direct input & FK masks only \\
Robot mesh & Geometry prior & No & No & Foreground mask \\
Source-time stereo depth & No & Scale only$^\dagger$ & No & Scale only$^\dagger$ \\
Right-eye RGB/depth maps & Evaluation only & Evaluation only & Evaluation only & Evaluation only \\
3D track references & Evaluation only & Evaluation only & Evaluation only & Evaluation only \\
Post-hoc target-depth fitting & No & No & No & No \\
\hline
\end{tabular}

\par\smallskip
\begin{minipage}{\linewidth}\footnotesize
$^\dagger$~Scalars use recorded left-eye stereo depth before inference or optimization, including for rollout inputs. They carry indirect information from the recorded stereo pairs, but no right-eye image or depth map is supplied directly.\\
$^\ddagger$~All methods receive calibrated camera poses, including the wrist trajectory. SoM uses robot states to construct foreground masks, not as explicit motion-conditioning inputs.
\end{minipage}
\end{table}

Recorded reconstruction uses real observations; rollout reconstruction uses
generated observations with the same calibration, trajectories, queries,
and recorded targets.
Rollout scores therefore combine generation and reconstruction errors rather
than measuring reproduction of generated inputs alone.
The supplied robot trajectory is unchanged, so similar robot EPE across
settings does not establish accurate visual motion generation.
For our model, paired-right observations provide training supervision on
training episodes, but never reconstruction tokens or recurrent feedback on
evaluation clips.
These are system-level comparisons: frame counts, auxiliary inputs, and
training differ.
Appendix~\ref{app:latent_interface} describes the optional latent-token interface.

\subsection{Baseline adaptations and coverage}
\label{app:baseline_protocols}

MoVieS, 4DGT, and Shape of Motion follow their released monocular settings:
each independently reconstructs the wrist and two external-camera streams.
We express their three fields in the common metric robot base frame,
concatenate their Gaussians, and render all six evaluation cameras.
This \emph{camera-wise reconstruction followed by calibrated merging} differs
from our joint field formation.
Metric-scale conventions are method-specific; no main-comparison field is
fitted to target-view depth after reconstruction.

\paragraph{MoVieS.}
Released MoVieS~\citep{lin2026movies} uses the five designated source times
per camera.
To stabilize scale for fixed external cameras and short wrist trajectories,
we compute one normalization scalar per camera from recorded left-eye stereo
depth at these times, using the mean distance of valid back-projected
source points.
Camera translations are normalized before inference, and the same factor
restores the predicted field to metres afterwards.
The three fields are rigidly transformed into the robot base frame and
concatenated for source- and novel-view rendering.
Scale estimation uses neither depth outside the five designated times nor
right-eye depth maps; a query coinciding with an input time does not introduce
additional depth access.

\paragraph{4DGT.}
We use the public first-stage 4DGT~\citep{xu20254dgt} checkpoint with all
17 frames per camera.
Inference images are resized to $182\times322$ with corresponding intrinsics
rescaling.
Following the released pipeline, camera poses are rigidly recentered around
the mean input camera pose.
This changes the coordinate frame without rescaling the supplied metric
translations.
After inference, we undo recentering and concatenate the three fields in the
robot base frame for rendering.
Expected camera-$z$ depth is scored in metres, without stereo-depth input or
output-scale fitting to evaluation targets.
Appendix~\ref{app:frame_count} uses the same protocol with five input frames.

\paragraph{Shape of Motion.}
Released Shape of Motion~\citep{wang2025shapeofmotion} uses all 17 frames
independently per camera.
Preprocessing uses UniDepth-v1 metric depth, Depth-Anything relative depth,
and BootsTAPIR tracks.
To correct camera-specific depth-scale biases before optimization, we compute
\begin{equation}
\gamma_v = \operatorname{median}_{s\in\mathcal S,\,\mathbf u\in\mathcal P^{\mathrm{st}}_{v,s}} \frac{D^{\mathrm{stereo}}_{v,s}(\mathbf u)}{D^{\mathrm{UniDepth}}_{v,s}(\mathbf u)},
\label{eq:som_scale_anchor}
\end{equation}
where $\mathcal S$ contains the five designated source times and
$\mathcal P^{\mathrm{st}}_{v,s}$ is the valid cached stereo-pixel support.
UniDepth estimates are multiplied by $\gamma_v$ before subsequent
preprocessing and optimization; recorded stereo depth is not used thereafter.
Each camera uses a single $\gamma_v$ across all 17 frames, rather than separate
scales for individual times or evaluation views.

Calibrated cameras are used directly, scene renormalization is disabled,
and projected FK robot masks replace interactive foreground selection.
All runs share the resolution-adjusted settings of 3-pixel erosion and
6-pixel dilation.
Each camera-specific scene follows the released 500-epoch schedule, giving
1,500 updates for 17 frames with batch size eight.
Learned camera corrections are not used for evaluation.
The three dynamic fields are combined in the calibrated robot base frame and
rendered from all six evaluation cameras.

\paragraph{Dynamic displacement output.}
Persistent Gaussian identities define anchor-to-query displacement for
RoGSW4RLD and MoVieS.
The evaluation wrapper obtains 4DGT positions from its released per-Gaussian
position polynomial and SoM positions from its optimized motion bases.
Each method's displacements are rendered through the anchor cameras and
passed to the same track-EPE scorer.

\paragraph{Coverage and exclusions.}
RoGSW4RLD, MoVieS, and 4DGT cover all 256 selected clips.
The MoVieS world-coordinate render-parity check is an audit, not a
cohort-selection criterion.
Near-threshold cases remain included after camera-point consistency checks;
their effect on reported means is below the displayed precision.

SoM is evaluated on 246 clips per setting.
A clip is excluded only when at least one of its three camera optimizations
fails under the common foreground-mask settings, preventing merged-field
construction.
Its reported means therefore condition on successful runs rather than the
full cohort.

Undefined 4DGT depth pixels are assigned zero and remain penalized inside the
fixed reference mask.
Robot EPE uses each method row's evaluated clips, with the required track
sidecars available; empty track sets are handled as specified below.

\subsection{Metric definitions}
\label{app:metric_definitions}

\paragraph{Image metrics.}
PSNR uses full-image RGB MSE in $[0,1]$.
SSIM~\citep{wang2004ssim} uses an $11\times11$ Gaussian window with
$\sigma=1.5$; LPIPS~\citep{zhang2018lpips} uses VGG with RGB normalized to
$[-1,1]$.
Source- and novel-view scores use their respective three-camera groups.
PSNR is computed per image before averaging, not from pooled multi-camera MSE.
Scoring covers the full target image, including background and regions
occluded in reconstruction inputs.

\paragraph{Metric depth.}
For each camera and query time, let $\hat z_{\mathbf u}$ denote rendered
expected camera-$z$ depth and $z_{\mathbf u}$ the cached
FoundationStereo~\citep{wen2025foundationstereo} reference, both in metres.
On fixed valid pixel support $\mathcal P$, we compute
\begin{align}
\operatorname{AbsRel} &= \frac{1}{|\mathcal P|}\sum_{\mathbf u\in\mathcal P}\frac{|\hat z_{\mathbf u}-z_{\mathbf u}|}{z_{\mathbf u}}, \\
\operatorname{RMSE} &= \sqrt{\frac{1}{|\mathcal P|}\sum_{\mathbf u\in\mathcal P}(\hat z_{\mathbf u}-z_{\mathbf u})^2}, \\
\delta_1 &= \frac{1}{|\mathcal P|}\sum_{\mathbf u\in\mathcal P}\mathbf 1\!\left[\frac{z_{\mathbf u}}{1.25}<\hat z_{\mathbf u}<1.25z_{\mathbf u}\right].
\end{align}
The interval form of $\delta_1$ implements the usual ratio threshold for
positive reference depths while rejecting non-positive predictions.

Source-view support requires finite reference depth with $0.1<z<5$~m.
Right-eye support follows the cached validity mask and its configured metric
range.
Neither depends on predicted opacity, depth, or error.
Non-positive predictions remain in AbsRel and RMSE.
Predictions are neither clipped to the reference range nor rescaled to target
depth in the main evaluation.
These metrics measure agreement with stereo estimates, not sensor-measured
ground truth.

\paragraph{Robot displacement and placement.}
For evaluation notation, let $\boldsymbol\mu_i^{\mathrm{base}}(t)$ denote
Gaussian centers expressed in metres in the robot base frame, related to the
reconstruction frame by the inverse fixed transform and unit conversion in
Appendix~\ref{app:coordinates}.
Anchor-to-query displacement is
$\boldsymbol\mu_i^{\mathrm{base}}(t)-\boldsymbol\mu_i^{\mathrm{base}}(a)$.
Rendering these displacements into anchor camera $v$ and normalizing by
accumulated opacity gives $\hat{\mathbf d}_{v,a\to t}$.
For valid robot tracks $\mathcal T_{\mathrm{robot}}^{a,t}$, with anchor
camera $v_j$ and pixel $\mathbf u_j^a$, robot EPE is
\begin{equation}
e_{\mathrm{track}}(a,t)=\frac{1000}{|\mathcal T_{\mathrm{robot}}^{a,t}|}\sum_{j\in\mathcal T_{\mathrm{robot}}^{a,t}}\left\|\hat{\mathbf d}_{v_j,a\to t}(\mathbf u_j^a)-(\mathbf X_j^t-\mathbf X_j^a)\right\|_2\quad\text{(mm)}.
\end{equation}
Here $\mathbf X_j^a$ and $\mathbf X_j^t$ are cached track positions in the
same metric frame.
Robot membership uses fixed FK-mesh support; anchor-time comparisons and
empty track sets are omitted.
EPE measures displacement, not absolute endpoint location or agreement with
analytic FK.
Stage~2 preserves individual Gaussian displacements, but rendered EPE can
change through compositing weights.

Robot depth error is $1000\,\operatorname{mean}|\hat z-z|$ (mm) on the
reference robot core, pooling valid source-camera pixels.
The core requires a common link throughout a $3\times3$ neighborhood, valid
stereo depth, and stereo--mesh depth consistency under the fixed threshold.
These stereo checks define evaluation support only.
The metric measures supported robot-surface placement, not joint-center or
end-effector localization.
Non-robot mover EPE is omitted because manual auditing identified false
static-object drift in cached labels.

\paragraph{Aggregation and uncertainty.}
Image metrics are averaged over cameras and distinct query times within
each view group, then equally over evaluated clips.
Each clip corresponds to one episode, so episodes rather than scenes
receive equal weight.
Robot metrics pool valid support at each query time before averaging over
times and clips.
We report 95\% confidence intervals from 10,000 clip-level bootstrap
resamples per method.
They describe test-clip sampling variability, not variability across
training seeds.
Table~\ref{tab:main_uncertainty} reports the central comparison.


\section{Additional Baseline Comparisons and Uncertainty}
\label{app:extended_results}
\label{app:rollout_consistency}

\subsection{Qualitative reconstruction from recorded observations}
\label{app:recorded_qualitative}

\begin{figure}[tbp]
    \centering
    \includegraphics[width=\linewidth]{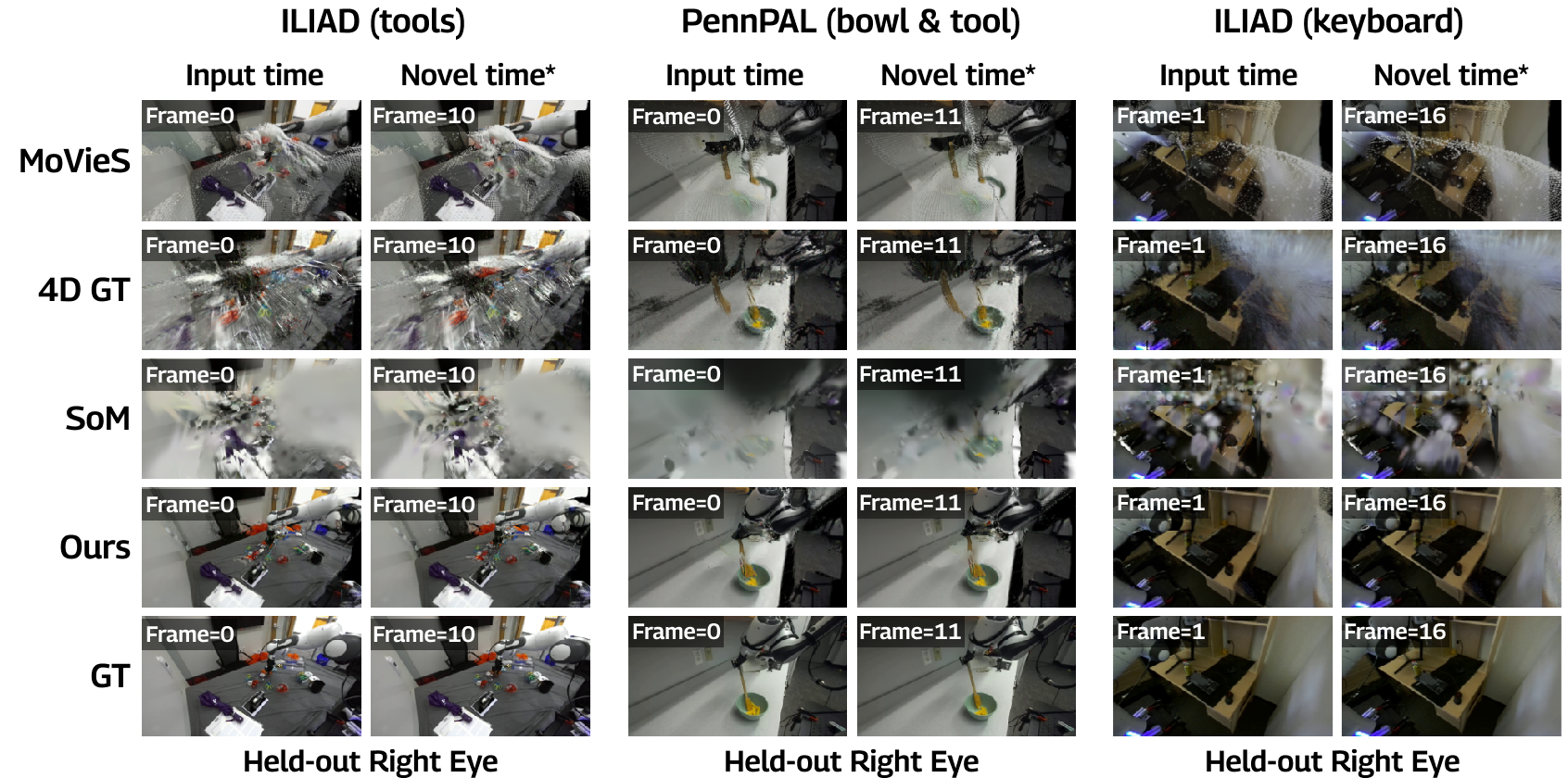}
    \caption{\textbf{Novel-view reconstruction from recorded observations.}
    RoGSW4RLD shows less ghosting and clearer scene details at held-out right-eye cameras. The bottom row shows recorded references; asterisks mark non-input times for RoGSW4RLD and MoVieS only.}
    \label{fig:recorded_qualitative}
    \end{figure}

Figure~\ref{fig:recorded_qualitative} compares reconstructions from recorded
inputs at held-out right-eye cameras.
These examples show fewer ghosting artifacts and clearer object boundaries
with RoGSW4RLD, complementing the generated-rollout source-view comparison
in Figure~\ref{fig:cosmos_qualitative}.

\subsection{Metrics omitted from the main table}
\label{app:additional_metrics}

Table~\ref{tab:additional_metrics} complements
Table~\ref{tab:reconstruction_main} using identical predictions, evaluated
clips, reference supports, and metric conventions.
Most depth gains are already present in Stage~1; Stage~2 further improves
source-view SSIM, with smaller changes in depth RMSE and $\delta_1$.

\begin{table}[!ht]
\centering
\caption{Complementary reconstruction metrics. Means use 256 clips for MoVieS, 4DGT, and our model, and 246 successful clips for SoM. $^\dagger$: one source-stereo scale scalar per camera before reconstruction.}
\label{tab:additional_metrics}
\label{tab:recorded_rendering_full}
\label{tab:recorded_geometry_full}
\label{tab:rollout_rendering_full}
\label{tab:rollout_geometry_full}
\small
\setlength{\tabcolsep}{8pt}
\renewcommand{\arraystretch}{1.1}
\begin{tabular}{@{}lccc@{}}
\hline
Method & \shortstack{Source\\SSIM $\uparrow$} & \shortstack{Source\\RMSE (m) $\downarrow$} & \shortstack{Novel\\$\delta_1$ (\%) $\uparrow$} \\
\hline
\multicolumn{4}{@{}l}{\textit{Recorded observations}} \\
MoVieS$^\dagger$ & 0.713 & 0.333 & 55.1 \\
4DGT (17f) & 0.585 & 0.414 & 31.1 \\
SoM$^\dagger$ & 0.573 & 0.473 & 38.1 \\
Ours (Stage~1) & 0.817 & 0.257 & 77.2 \\
Ours (Stage~1+2) & 0.866 & 0.255 & 77.5 \\
\hline
\multicolumn{4}{@{}l}{\textit{Cosmos~3 rollouts}} \\
MoVieS$^\dagger$ & 0.639 & 0.354 & 52.0 \\
4DGT (17f) & 0.533 & 0.427 & 30.6 \\
SoM$^\dagger$ & 0.531 & 0.473 & 37.8 \\
Ours (Stage~1) & 0.736 & 0.286 & 73.0 \\
Ours (Stage~1+2) & 0.775 & 0.284 & 73.2 \\
\hline
\end{tabular}
\end{table}

\subsection{Confidence intervals}
\label{app:confidence_intervals}

Table~\ref{tab:main_uncertainty} reports uncertainty across evaluated clips,
not training seeds.
SoM intervals condition on successful runs.
These are intervals for individual method means, not paired differences;
they do not directly quantify uncertainty in the incremental effect of Stage~2.

\begin{table}[!ht]
\centering
\caption{Main-comparison means and 95\% clip-bootstrap confidence intervals from 10,000 resamples. MoVieS, 4DGT, and our model use 256 clips; SoM uses 246. $^\dagger$: one source-stereo scale scalar per camera.}
\label{tab:main_uncertainty}
\footnotesize
\setlength{\tabcolsep}{5pt}
\renewcommand{\arraystretch}{1.13}
\begin{tabular}{@{}lccc@{}}
\hline
Method & Novel PSNR (dB) $\uparrow$ & Novel AbsRel $\downarrow$ & Robot EPE (mm) $\downarrow$ \\
\hline
\multicolumn{4}{@{}l}{\textit{Recorded observations}} \\
MoVieS$^\dagger$ & 13.93 [13.77, 14.09] & 0.373 [0.356, 0.390] & 22.2 [20.5, 24.2] \\
4DGT (17f) & 12.90 [12.75, 13.05] & 0.681 [0.659, 0.704] & 31.0 [28.8, 33.2] \\
SoM$^\dagger$ & 12.14 [12.00, 12.28] & 0.429 [0.413, 0.445] & 32.2 [30.1, 34.4] \\
Ours (Stage~1) & 15.52 [15.35, 15.70] & 0.200 [0.193, 0.208] & 8.6 [7.8, 9.6] \\
Ours (Stage~1+2) & 16.08 [15.90, 16.27] & 0.197 [0.190, 0.205] & 8.6 [7.7, 9.6] \\
\hline
\multicolumn{4}{@{}l}{\textit{Cosmos~3 rollouts}} \\
MoVieS$^\dagger$ & 13.69 [13.53, 13.85] & 0.404 [0.385, 0.424] & 22.6 [20.9, 24.5] \\
4DGT (17f) & 12.78 [12.64, 12.93] & 0.697 [0.674, 0.720] & 32.3 [30.0, 34.7] \\
SoM$^\dagger$ & 12.05 [11.91, 12.19] & 0.440 [0.423, 0.458] & 37.6 [35.0, 40.4] \\
Ours (Stage~1) & 15.04 [14.87, 15.22] & 0.218 [0.209, 0.227] & 8.7 [7.8, 9.7] \\
Ours (Stage~1+2) & 15.52 [15.34, 15.71] & 0.215 [0.206, 0.224] & 8.7 [7.8, 9.7] \\
\hline
\end{tabular}
\end{table}

\subsection{Number of input frames}
\label{app:frame_count}

We compare 4DGT using all 17 frames per camera with the same five source
times supplied to our model.
Both variants retain camera-wise reconstruction followed by calibrated
merging, using the same checkpoint, input resolution, query plan, evaluation
cameras, and recorded references.
This measures input-density sensitivity within 4DGT rather than isolating
architectural differences between methods.

\begin{table}[!ht]
\centering
\caption{4DGT with five or 17 input frames per camera under the same camera-wise reconstruction and calibrated-merging protocol on 256 clips.}
\label{tab:4dgt_frame_count}
\small
\setlength{\tabcolsep}{5pt}
\renewcommand{\arraystretch}{1.1}
\begin{tabular}{@{}lccccc@{}}
\hline
& & Source & \multicolumn{3}{c}{Novel view} \\
Setting & Frames & PSNR $\uparrow$ & PSNR $\uparrow$ & LPIPS $\downarrow$ & AbsRel $\downarrow$ \\
\hline
Recorded & 5 & 15.63 & 12.78 & 0.603 & 0.803 \\
Recorded & 17 & 16.54 & 12.90 & 0.584 & 0.681 \\
Cosmos~3 & 5 & 15.20 & 12.69 & 0.608 & 0.815 \\
Cosmos~3 & 17 & 15.78 & 12.78 & 0.591 & 0.697 \\
\hline
\end{tabular}
\end{table}

Seventeen frames improve every reported metric in both settings, but do not
close the gap to our model.
The main comparison therefore retains this stronger 4DGT configuration rather
than restricting its temporal input to match ours.

\subsection{Scale-adjusted depth diagnostic}
\label{app:scale_diagnostic}

The main evaluation scores metric reconstructions without target-view scale
fitting.
Here, we rescore 4DGT and SoM after multiplying each predicted depth image by
its median reference-to-prediction depth ratio.
A separate factor is fitted for each camera and query time, rather than one
global factor for the field.
This target-assisted operation changes only scored depth images; it neither
rescales nor rerenders the Gaussian field.
Unadjusted scores appear in Table~\ref{tab:reconstruction_main}.

\begin{table}[!ht]
\centering
\caption{Target-assisted median-scale depth diagnostic (AbsRel $\downarrow$). $^\dagger$: source-stereo scaling before reconstruction. $^\ddagger$: per-image target-assisted scaling used only here, not in the main comparison.}
\label{tab:scale_diagnostic}
\small
\setlength{\tabcolsep}{9pt}
\renewcommand{\arraystretch}{1.1}
\begin{tabular}{@{}lcccc@{}}
\hline
& \multicolumn{2}{c}{Recorded} & \multicolumn{2}{c}{Cosmos~3 rollouts} \\
Method & Source & Novel & Source & Novel \\
\hline
4DGT$^\ddagger$ & 0.320 & 0.411 & 0.338 & 0.417 \\
SoM$^{\dagger\ddagger}$ & 0.402 & 0.489 & 0.411 & 0.498 \\
\hline
\end{tabular}
\end{table}

For 4DGT, novel-view AbsRel decreases from 0.681 to 0.411 on recorded inputs
and from 0.697 to 0.417 on rollouts.
SoM instead worsens from 0.429 to 0.489 and from 0.440 to 0.498.
Thus, this correction substantially reduces 4DGT's error but leaves a sizable
residual; it does not improve SoM.
The diagnostic does not establish a single field-level scale correction.

\section{Additional Design Analyses}
\label{app:ablations}

\subsection{Intervention definitions and Stage 1 effects}
\label{app:joint_ablation}
\label{app:intervention_details}

Table~\ref{tab:controlled_ablation} examines joint lifting and refinement in
panel~(a), and robot geometry and motion in panel~(b).
We define these frozen-checkpoint interventions and test whether the
robot-specific effects precede Stage~2.

The cross-camera intervention blocks attention to other cameras in Stage~1
while retaining within-camera temporal interaction, calibration, and
robot inputs.
Camera-specific fields are then combined by calibrated merging.
This probes the jointly trained checkpoint's dependence on inference-time
cross-camera exchange, not the performance of separately trained monocular
models.

Disabling FK conditioning suppresses the kinematic input to learned
deformation while retaining mesh anchoring.
Disabling mesh anchoring restores learned source geometry for robot-supported
pixels and removes the associated mesh anchoring pathway while retaining
FK conditioning.
This is a pathway-level intervention, not an isolated depth replacement.
Calibration, including FK-derived wrist poses, remains unchanged.
Conditioning dropout supports FK removal; mesh anchoring removal and removal of
both are out-of-distribution diagnostics.
No variant is retrained.

\begin{table}[!ht]
\centering
\caption{Robot-specific interventions in Stage~1 on 256 recorded clips. Table~\ref{tab:controlled_ablation} reports the corresponding Stage~1+2 interventions for recorded observations and Cosmos~3 rollouts.}
\label{tab:stagewise_kinematics}
\small
\setlength{\tabcolsep}{10pt}
\renewcommand{\arraystretch}{1.1}
\begin{tabular}{@{}lcc@{}}
\hline
Configuration & \shortstack{Robot EPE\\(mm) $\downarrow$} & \shortstack{Robot depth\\error (mm) $\downarrow$} \\
\hline
Full Stage~1 & 8.64 & 13.4 \\
w/o FK conditioning & 30.14 & 21.5 \\
w/o mesh anchoring & 8.80 & 36.5 \\
w/o both & 30.11 & 41.6 \\
\hline
\end{tabular}
\end{table}

The complementary pattern already appears before refinement: FK removal raises
displacement EPE from 8.64 to 30.14~mm, whereas mesh anchoring removal raises
robot depth error from 13.4 to 36.5~mm with EPE at 8.80~mm.
Stage~2 therefore does not create this separation between placement and
displacement.

\subsection{Learned motion and analytic FK transport}
\label{app:rigid_policy}

We replace learned robot-motion prediction with analytic FK transport of
mesh-anchored source points and their covariances along the supplied link
trajectories.
Inputs and learned weights remain unchanged.
This geometry-transport diagnostic evaluates displacement and placement against
the recorded references, not prediction of unknown robot actions.
Because both centers and covariances change, it is not an isolated center-motion
ablation.

\begin{table}[!ht]
\centering
\caption{Score changes from learned motion to analytic FK transport. Entries are analytic-FK minus learned-motion scores; positive PSNR and negative errors favor analytic transport.}
\label{tab:rigid_policy}
\small
\setlength{\tabcolsep}{5pt}
\renewcommand{\arraystretch}{1.1}
\begin{tabular}{@{}llrrrr@{}}
\hline
Input & Stage & \shortstack{$\Delta$ Novel\\PSNR (dB)} & $\Delta$ AbsRel & \shortstack{$\Delta$ Robot\\EPE (mm)} & \shortstack{$\Delta$ Robot depth\\error (mm)} \\
\hline
Recorded & Stage~1 & 0.00 & -0.004 & -0.16 & +1.7 \\
Recorded & Stage~1+2 & +0.01 & -0.004 & -0.18 & +0.9 \\
Cosmos~3 & Stage~1+2 & +0.01 & -0.004 & -0.20 & +0.9 \\
\hline
\end{tabular}
\end{table}

Analytic transport lowers robot EPE by 0.16--0.20~mm but increases robot depth
error by 0.9--1.7~mm; novel-view PSNR changes by at most 0.01~dB.
These results do not support superior robot-displacement accuracy from
learned deformation.
Its role is instead to maintain one deformation parameterization for robot
and non-robot dynamics.
The small score differences describe this diagnostic, not a demonstrated
statistical advantage.

\subsection{Number of rendering-feedback refinement updates}
\label{app:recurrence}

Table~\ref{tab:recurrence_sweep} varies the number of updates $r$ using the
same Stage~2 checkpoint and initial Stage~1 fields.
The fixed 32-clip subset evaluates the quality--computation trade-off;
its refinement-only timings are distinct from the complete-pipeline
measurements in Table~\ref{tab:reconstruction_main}.

\begin{table}[!ht]
\centering
\caption{Refinement sweep on the first 32 evaluation clips under the fixed-window protocol. Quality metrics use all 32 clips. Refinement time uses CUDA-synchronized measurements on 27 clips after excluding each process's warm-up clip; lifting and final query rendering are excluded.}
\label{tab:recurrence_sweep}
\small
\setlength{\tabcolsep}{5pt}
\renewcommand{\arraystretch}{1.1}
\begin{tabular}{@{}ccccccc@{}}
\hline
Updates $r$ & \shortstack{Source\\PSNR $\uparrow$} & \shortstack{Novel\\PSNR $\uparrow$} & \shortstack{Source\\LPIPS $\downarrow$} & \shortstack{Source\\AbsRel $\downarrow$} & \shortstack{Robot EPE\\(mm) $\downarrow$} & \shortstack{Time (s)\\$\downarrow$} \\
\hline
0 & 22.59 & 16.02 & 0.182 & 0.178 & 6.21 & -- \\
1 & 24.94 & 16.39 & 0.169 & 0.177 & 6.22 & 2.61 \\
2 & 25.59 & 16.44 & 0.168 & 0.177 & 6.22 & 4.78 \\
3 & 25.80 & 16.47 & 0.171 & 0.176 & 6.22 & 6.95 \\
4 & 25.87 & 16.48 & 0.173 & 0.176 & 6.22 & 9.12 \\
\hline
\end{tabular}
\end{table}

Most PSNR gains occur within two updates, with diminishing gains thereafter.
Source LPIPS is lowest at two updates and then worsens slightly, so additional
updates do not uniformly improve image quality.
Source AbsRel and robot EPE change little on this subset.

The time-shared correction
$\boldsymbol\mu_i^{(r)}(t)=\boldsymbol\mu_i^{(0)}(t)+\boldsymbol\delta_i^{(r)}$
translates each Gaussian's entire Stage~1 center trajectory, preserving its
displacement between query times exactly.
This guarantee does not extend to rendered depth or EPE: changes in placement,
covariance, opacity, and visibility alter compositing weights.
Stage~2 therefore refines rendering without independently re-estimating
center trajectories.

\subsection{Reconstruction and rendering cost}
\label{app:efficiency}

Table~\ref{tab:reconstruction_main} reports reconstruction and query-rendering
time and peak allocated GPU memory.
Below we specify their method-dependent measurement scopes.
Model loading and world-model generation are excluded.

\paragraph{Our model.}
Stage~1 requires 0.436~s for joint lifting and 0.153~s for query rendering,
totaling 0.59~s after rounding.
The full model uses 0.436~s for lifting, 8.417~s for four refinement updates,
and 0.228~s for final rendering, totaling 9.08~s.
Both stages use fixed inference weights without per-scene optimization.

\paragraph{MoVieS.}
Timing includes three camera-wise inferences, transformation into the common
metric robot frame, Gaussian merging, and rendering of the common query set.
Diagnostic per-view renders are excluded.

\paragraph{4DGT.}
The 17-frame workload comprises 0.09~s of preprocessing, 1.93~s of camera-wise
inference, 2.59~s of query rendering, and 0.19~s of displacement rendering,
totaling 4.80~s.
The five-frame variant is timed separately.

\paragraph{Shape of Motion.}
Timing includes preprocessing and optimization for all three camera streams
plus merged-field rendering.
It sums the camera-run workloads rather than reporting rendering latency from
an already optimized scene.

\paragraph{Measurement conditions.}
RoGSW4RLD, MoVieS, and 4DGT use otherwise idle B200 workloads with explicit
CUDA synchronization.
SoM is measured under concurrent workloads, so its aggregate run time is not
an isolated single-GPU end-to-end latency.

Peak allocation is measured using \texttt{torch.cuda.max\_memory\_allocated}
after resetting the peak statistic for each clip.
Reported memory includes model weights but excludes the CUDA context and
unused allocator cache.

RoGSW4RLD and MoVieS share a process holding multiple models.
Their method-specific estimates add the invoked modules' parameter and buffer
memory to the peak allocation increase above the pre-invocation baseline,
averaged over 16 clips after warm-up.
These are not unadjusted shared-process peaks.
4DGT and SoM use method-specific processes and aggregate per-clip peaks over
evaluated clips.
SoM uses the largest peak among its three camera optimizations and excludes
merged-field rendering memory.
These differing scopes should be retained when comparing memory requirements.
Network-training memory is not reported.


\section{Optional Latent-Token Interface}
\label{app:latent_interface}

\subsection{Adapter and training}
\label{app:adapter_training}

Following the latent embedding design of
Diff4Splat~\citep{pan2026diff4splat}, the adapter maps each camera region of
the Cosmos~3~\citep{nvidia2026cosmos3} latent to the Stage~1 view-time
token grid.
Latents have 48 channels, spatial stride eight, and temporal stride four.
Cross-camera interaction remains in the joint Stage~1 backbone.

Adapter training first aligns outputs with
DINOv2~\citep{oquab2024dinov2} visual tokens, then distills intermediate
Stage~1 features and log-depth predictions.
Both reconstruction stages remain frozen.
Starting from the token-aligned checkpoint, the second phase uses 300 AdamW
updates with learning rate $5\times10^{-5}$ and weight decay 0.01.
Feature and log-depth losses each have weight 1.0; the input-token loss has
weight 0.1.
Decoded RGB remains in the Gaussian-attribute shortcut and the refinement
observation and error encoders, so this interface is not RGB-free.

\subsection{Effect of replacing RGB tokens}
\label{app:rgb_token_effect}

Both interfaces share downstream reconstruction weights, calibrated cameras,
robot inputs, and the evaluation protocol.
Table~\ref{tab:input_interface} reports the latent-token path; RGB references
appear in Table~\ref{tab:reconstruction_main}, with robot depth error in
Table~\ref{tab:controlled_ablation}(b).

\begin{table}[!ht]
\centering
\caption{Optional latent-token interface with Stage~1+2 and four refinement updates on 256 clips. Robot EPE measures track displacement; robot depth error measures source-view placement on the reference robot core. Cosmos~3 rollouts supply generated latents directly.}
\label{tab:input_interface}
\small
\setlength{\tabcolsep}{8pt}
\renewcommand{\arraystretch}{1.1}
\begin{tabular}{@{}lcccc@{}}
\hline
Input & \shortstack{Novel PSNR\\(dB) $\uparrow$} & \shortstack{Novel\\AbsRel $\downarrow$} & \shortstack{Robot EPE\\(mm) $\downarrow$} & \shortstack{Robot depth\\error (mm) $\downarrow$} \\
\hline
Recorded & 15.32 & 0.293 & 9.0 & 20.7 \\
Cosmos~3 & 14.98 & 0.298 & 9.0 & 20.7 \\
\hline
\end{tabular}
\end{table}

Relative to RGB, the latent interface lowers novel-view PSNR by 0.76~dB and
raises AbsRel by 0.096 on recorded inputs; the rollout gaps are 0.54~dB and
0.083.
Robot EPE increases by approximately 0.4 and 0.3~mm at the reported precision,
while robot depth error increases by 7.0 and 5.2~mm, respectively.
These relatively small EPE changes occur with mesh anchoring and
FK conditioning retained; they do not establish motion preservation by
latent tokens alone.

\subsection{Spatial-resolution diagnostic}
\label{app:spatial-resolution-diagnostics}
We compare the latent interface with RGB downsampled by a factor of eight
per dimension and restored before reconstruction.
This probes sensitivity to spatial detail, not equivalence between latent
compression and image downsampling.

\begin{table}[!ht]
\centering
\caption{Stage~1-only spatial-resolution diagnostic on the first 16 evaluation clips under the fixed-window protocol. Metrics use source views; ``Far'' restricts AbsRel to reference camera-$z$ depths between 1.5 and 5~m. Every input frame and camera is area-downsampled and bilinearly restored in the $8\times$ RGB variant.}
\label{tab:latent_resolution}
\small
\setlength{\tabcolsep}{6pt}
\renewcommand{\arraystretch}{1.1}
\begin{tabular}{@{}lcccc@{}}
\hline
Input & PSNR (dB) $\uparrow$ & AbsRel $\downarrow$ & $\delta_1$ (\%) $\uparrow$ & Far AbsRel $\downarrow$ \\
\hline
Original RGB & 23.35 & 0.160 & 80.3 & 0.191 \\
$8\times$ down/up RGB & 20.76 & 0.227 & 63.3 & 0.482 \\
Latent adapter & 22.37 & 0.228 & 62.5 & 0.411 \\
\hline
\end{tabular}
\end{table}

Both alternatives increase overall depth AbsRel from 0.160 to 0.227--0.228.
However, latent tokens yield higher PSNR and lower far-depth error than
degraded RGB.
This is consistent with sensitivity to spatial detail, but does not isolate
spatial resolution from representation or adapter effects as the cause of
the latent gap.

\section{Scope and Limitations}
\label{app:limitations}
\label{app:qualitative}

The lab holdout evaluates transfer to ILIAD and PennPAL, whose episodes
are excluded from our training and validation of Stage~1, Stage~2,
and the optional adapter.
Evaluation uses calibrated wrist and external cameras, known robot geometry,
supplied trajectories, and 17-frame intervals.
Novel-view evaluation uses paired right-eye cameras; wider baselines,
new robot morphologies, and longer rollouts remain untested.
Errors in stereo and track references affect measured accuracy, and robot
EPE does not establish non-robot motion accuracy.

Baseline comparisons are system-level because training, input frame counts,
and auxiliary inputs differ.
MoVieS and SoM receive one scale scalar per camera estimated from recorded
source stereo depth; RoGSW4RLD and 4DGT receive no test-time stereo depth.
SoM uses projected FK foreground masks and completes 246 of 256 clips,
so its reported means are conditional on successful runs.

Cosmos~3 scores compare reconstructions of generated observations against
recorded futures, combining generation and reconstruction errors.
Supplied robot trajectories remain unchanged between recorded and rollout
settings, so stable robot EPE does not demonstrate accurate visual
motion generation.

Reconstruction is also sensitive to the spatial detail available in its
visual inputs.
The 16-clip Stage~1 spatial-resolution diagnostic in Appendix~\ref{app:spatial-resolution-diagnostics} shows
that downsampling RGB inputs and resizing them back to the original
resolution degrades rendering and depth accuracy.
This supports sensitivity to input detail, but does not isolate rollout
resolution as the cause of the performance gap between recorded and
generated inputs.
Because RoGSW4RLD reconstructs pretrained world-model outputs rather than
learning a new transition model, it may benefit from future improvements
in the resolution and consistency of action-conditioned rollouts.

Stage~2 preserves each Gaussian's center displacement between query times
and therefore cannot correct displacement errors inherited from Stage~1.
These experiments evaluate reconstruction, not physical reasoning,
task success, or closed-loop robot control.

\end{document}